\documentclass{article}

\PassOptionsToPackage{numbers, sort&compress}{natbib}

\usepackage[preprint]{style/neurips_2019}

\usepackage{xspace}
\usepackage{graphicx}
\usepackage{float}
\usepackage{subcaption}

\usepackage{amsmath,amssymb,amsfonts}
\usepackage{algorithm}
\usepackage{algpseudocode}
\usepackage{graphicx}
\usepackage{textcomp}
\usepackage{xcolor}
\usepackage{tikz}
\usetikzlibrary{bayesnet}
\usetikzlibrary{calc,arrows.meta,positioning}
\usepackage{mathtools}
\usepackage{pgfplotstable}

\usepackage{rotating}

\newcommand*\real{{\rm I\!R}}
\newcommand*\normal{{\mathcal{N}}}

\newcommand*\E{\mathbb{E}}
\newcommand*\expected{\E}
\newcommand*\Var{\mathbb{V}}

\newcommand*\mixf{m}

\newcommand{\eq}{Eq.\xspace}

\newcommand{\acro}[1]{\textsc{#1}\xspace}

\newcommand{\dgp}{\acro{DGP}}

\newcommand{\cmgpaggr}{\acro{mr-gprn}}

\newcommand{\cmgpdgp}{\acro{mr-dgp}}
\newcommand{\cmgpgp}{\acro{mr-gp}}

\newcommand{\centerpoint}{\acro{center-point} }
\newcommand{\dgpcascade}{\acro{mr-cascade} }

\newcommand{\vbagg}{\acro{VBAgg-normal}}

\newcommand{\srcN}{\mathcal{A}}
\newcommand{\srcn}{a}
\newcommand{\uParent}{\text{Pa}}

\newcommand{\res}{\mathcal{S}}
\newcommand{\cweight}{\phi}

\newcommand{\epochs}{E}

\newcommand{\numP}{\mathcal{K}}
\newcommand{\nump}{k}

\newcommand{\parent}{\mathcal{P}}
\newcommand{\numL}{\mathcal{L}}

\DeclareMathOperator*{\argmax}{arg\,max} 
\DeclareMathOperator*{\argmin}{arg\,min}

\newcommand{\uSec}{Section\xspace}

\newcommand{\uTable}{Table\xspace}

\newcommand{\uFig}{Fig.\xspace}

\usepackage[utf8]{inputenc} 
\usepackage[T1]{fontenc}    
\usepackage{hyperref}       
\usepackage{url}            
\usepackage{booktabs}       
\usepackage{amsfonts}       
\usepackage{nicefrac}       
\usepackage{microtype}      
\usepackage{multirow}

\usepackage{pgfplots}
\inputencoding{latin2}
\inputencoding{utf8}
\DeclareUnicodeCharacter{2212}{-}

\usepackage[toc,page]{appendix}

\definecolor{asparagus}{rgb}{0.53, 0.66, 0.42}

\title{Multi-resolution Multi-task Gaussian Processes}

\author{%
  Oliver Hamelijnck \\
  The Alan Turing Institute\\
  Department of Computer Science \\
  University of Warwick \\
  \texttt{ohamelijnck@turing.ac.uk}
   \And 
  Theodoros Damoulas \\
  The Alan Turing Institute \\
  Depts. of Computer Science \& Statistics\\
  University of Warwick \\
  \texttt{tdamoulas@turing.ac.uk}
    \And 
  Kangrui Wang \\
  The Alan Turing Institute\\
  Department of Statistics \\
  University of Warwick \\
  \texttt{kwang@turing.ac.uk}
   \And 
  Mark Girolami \\
  The Alan Turing Institute \\
  Department of Engineering \\
  University of Cambridge\\
  \texttt{mgirolami@turing.ac.uk}
}

\begin{document}

\maketitle


\begin{abstract}
    We consider evidence integration from potentially dependent observation processes under varying spatio-temporal sampling resolutions and noise levels. We offer a multi-resolution multi-task (\textsc{MRGP}) framework that allows for both \emph{inter-task} and \emph{intra-task} multi-resolution and multi-fidelity. We develop shallow Gaussian Process (GP) mixtures that approximate the difficult to estimate joint likelihood with a composite one and deep GP constructions that naturally handle biases. In doing so, we generalize existing approaches and offer information-theoretic corrections and efficient variational approximations. We demonstrate the competitiveness of MRGPs on synthetic settings and on the challenging problem of hyper-local estimation of air pollution levels across London from multiple sensing modalities operating at disparate spatio-temporal resolutions.
    

\end{abstract}

\section{Introduction}
The increased availability of ground and remote sensor networks coupled with new sensing modalities, arising from e.g. citizen science intiatives and mobile platforms, is creating new challenges for performing formal evidence integration. These multiple observation processes and sensing modalities can be dependent, with different signal-to-noise ratios and varying sampling resolutions across space and time. In our motivating application, London authorities measure air pollution from multiple sensor networks; high-fidelity ground sensors that provide frequent multi-pollutant readings, low fidelity diffusion tubes that only provide monthly single-pollutant readings, hourly satellite-derived information at large spatial scales, and high frequency medium-fidelity multi-pollutant sensor networks. Such a multi-sensor multi-resolution multi-task evidence integration setting is becoming prevalent across any real world application of machine learning.

The current state of the art, see also \uSec  \ref{sec:lit}, is assuming independent and unbiased observation processes and cannot handle the challenges of real world settings that are jointly \emph{non-stationary}, \emph{multi-task}, \emph{multi-fidelity}, and \emph{multi-resolution} \citep{Fox:2012:MGP:2999134.2999217,perdikaris2015multi,Perdikaris_Nonlinear_multi_fidelity,serban2017multiresolution,law2018variational,2018arXiv180902010S,pmlr-v71-adelsberg18a}. The latter challenge has recently attracted the interest of the machine learning community under the context of working with aggregate, binned observations \citep{2018arXiv180902010S,pmlr-v71-adelsberg18a,law2018variational} or the special case of natural language generation at multiple levels of abstraction \citep{serban2017multiresolution}. When the independence and unbiasedness assumptions are not satisfied they lead to posterior contraction, degradation of predictive performance and insufficient uncertainty quantification.

In this paper we introduce a multi-resolution multi-task GP framework that can integrate evidence from observation processes with varying support (e.g. partially overlapping in time and space), that can be dependent and biased while allowing for both \emph{inter-task} and \emph{intra-task} multi-resolution and multi-fidelity. Our first contribution is a shallow GP mixture, \cmgpaggr, that corrects for the dependency between observation processes through composite likelihoods and extends the Gaussian aggregation model of \citet{law2018variational}, the multi-task GP model of \citet{wilson_gprn}, and the variational lower bound of \citet{pmlr-v31-nguyen13b}. Our second contribution is a multi-resolution deep GP composition that can additionally handle biases in the observation processes and extends the deep GP models and variational lower bounds of \citet{pmlr-v31-damianou13a} and \citet{NIPS2017_7045} to varying support, multi-resolution data. Lastly, we demonstrate the superiority of our models on synthetic problems and on the challenging spatio-temporal setting of predicting air pollution in London at hyper-local resolution.

Sections \ref{sec:mr_gprn} and \ref{sec:mr_deep_gp} introduce our shallow GP mixtures and deep GP constructions respectively. In \uSec \ref{sec:exp} we demonstrate the empirical advantages of our framework versus the prior art followed by a additional related work in \uSec \ref{sec:lit} and our concluding remarks. Further analysis is provided in the Appendix with code available at \url{https://github.com/ohamelijnck/multi_res_gps}.

\section{Multi-resolution Multi-task Learning}
\label{sec:mres}

Consider $\srcN \in \mathbb{N}$ observation processes $\mathbf{Y}_a \in \real^{N_a \times P}$ across $P$ tasks with $N_a$ observations. Each process may be observed at varying resolutions that arises as the volume average over a sampling area $\mathcal{S}_a$. Typically we discretise the area $\mathcal{S}_a$ and so we overload $\mathcal{S}_a$ to denote these points. We construct $\srcN$ datasets $\{(\mathbf{X}_\srcn, \mathbf{Y}_\srcn) \}^{\srcN}_{\srcn=1}$, ordered by resolution size ($\mathbf{Y}_1$ is the highest, $\mathbf{Y}_\srcN$ is the lowest), where $\mathbf{X}_\srcn \in \real^{N_\srcn \times |\res_\srcn| \times D_\srcn}$ and $D_\srcn$ is the input dimension. For notational simplicity we assume that all tasks are observed across all observational processes, although this need not be the case.

In our motivating application there are multiple sensor networks (observation processes) measuring multiple air pollutants (tasks) such as CO$_2$, NO$_2$, PM$_{10}$, PM$_{2.5}$ at different sampling resolutions. These multi-resolution observations exist both within tasks, (\emph{intra-task multi-resolution}) when different sensor networks measure the same pollutant, and across tasks (\emph{inter-task multi-resolution}) when different sensor networks measure different but potentially correlated pollutants due to e.g. common emission sources. Our goal is to develop scalable, non-stationary non-parametric models for air pollution while delivering accurate estimation and uncertainty quantification.

%
%
\section{Multi-Resolution Gaussian Process Regression Networks (\cmgpaggr)}
\label{sec:mr_gprn}

 We first introduce a \emph{shallow} instantiation of the multi-resolution multi-task framework. \cmgpaggr is a shallow GP mixture, \uFig \ref{fig:mr_aggr_model_graphical_model}, that extends the Gaussian process regression network (GPRN) \cite{wilson_gprn}. Briefly, the GPRN jointly models all tasks by introducing $Q \in \mathbb{N}$ latent GPs that act as basis for the $P$ tasks. These GPs are combined using task specific weights, that are themselves GPs, resulting in $PQ \in \mathbb{N}$ latent weights $\mathbf{W_{p,q}}$. More formally, $\mathbf{f}_q \sim \mathcal{GP}(0, \mathbf{K}^{f}_{q})$, $\mathbf{W}_{p, q} \sim \mathcal{GP}(0, \mathbf{K}^{w}_{p, q})$ and each task $p$ is modelled as  $\mathbf{Y}_{p} = \sum^Q_{q=1} \mathbf{W}_{p,q} \odot \mathbf{f}_q + \epsilon_p$ where $\odot$ is the Hadamard product and $\epsilon \sim \normal(0, \sigma_p^2 \mathbf{I})$. The GPRN is an extension of the Linear Coregionalization Model (LCM) \cite{alvarez2012kernels} and can enable the learning of non-stationary processes through input dependent weights \cite{adams-stegle-2008a}.

%
%
\subsection{Model Specification}

We extend the GPRN model to handle multi-resolution observations by integrating the latent process over the sampling area for each observation. Apart from the standard inter-task dependency we would ideally want to be able to model additional dependencies between observation processes such as, for example, correlated noises. Directly modelling this additional dependency can quickly become intractable, due to the fact that it can vary in input space. If one ignores this dependency by assuming a product likelihood, as in \citep{law2018variational,Moreno-Munoz:2018:HMG:3327757.3327777},  then when violated the misspecification results in severe posterior contractions (see \uFig \ref{fig:mcmc_posterior_contraction}). To circumvent these extremes we approximate the full likelihood using a multi-resolution composite likelihood that corrects for the misspecification \cite{Varin11anoverview}. The posterior over the latent functions is now:
\begin{equation}
   p(\mathbf{W}, \mathbf{f} | \mathbf{Y}) \propto \underbrace{\prod^{\srcN}_{\srcn=1} \prod^{P}_{p=1} \prod^{N_a}_{n=1} \normal (\mathbf{Y}_{\srcn, p,n} | \frac{1}{|\res_{\srcn}|} \int_{\res_{\srcn, n}} \sum^{Q}_{q=1} \mathbf{W}_{p, q}(\mathbf{x}) \odot \mathbf{f}_q(\mathbf{x}) \mathop{d\mathbf{x}}, \sigma^2_{\srcn,p} \mathbf{I})^{\cweight}}_{\text{$\cmgpaggr$ Composite Likelihood}} \underbrace{\vphantom{\prod^{P_\srcn}_{p=1}} p(\mathbf{W}, \mathbf{f})}_{\text{GPRN Prior}}  \\
\label{eq:main_aggr_likelihood}
\end{equation}
 where $\cweight \in \real_{>0}$ are the composite weights that are critical for inference. The integral within the multi-resolution likelihood links the underlying latent process to each of resolutions; in general this is not available in closed form and so we approximate it by discretizing over a uniform grid. When we only have one task and $\mathbf{W}$ becomes a vector of constants we denote the model as $\cmgpgp$.
 
%

\begin{figure}[t]
\begin{minipage}[]{.5\textwidth}
      \centering
      \begin{tikzpicture}
         \node[latent] (y) {$\mathbf{Y}_{\srcn,p}$}; %
         \node[const, below=of y, outer sep=0.1cm] (ysigma) {$\sigma^2_{\srcn,p}$}; %
          \plate[] {plate4} {(ysigma)(y)} {$\srcN P$};%

         \node[latent, left=of y] (f) {$\mathbf{f}_q$}; %
         \plate[inner sep=0.25cm, xshift=-0.12cm, yshift=0.12cm] {platef} {(f)} {$Q$};%

         \node[latent, left=of y, below=of f, yshift=-0.2cm] (w) {$\mathbf{W}_{p,q}$}; %
         \plate[inner sep=0.25cm, xshift=-0.12cm, yshift=0.12cm] {plate2} {(w)} {$PQ$};%

         \node[obs, left=of f] (X) {$\mathbf{X}$}; %
         \node[obs, left=of w] (Xst) {$\mathbf{X}$}; %

         \edge {ysigma} {y} ; %
         \edge {f} {y} ; %
         \edge {w} {y} ; %
         \edge {X} {f} ; %
         \edge {Xst} {w} ; %
     \end{tikzpicture}     
\end{minipage}%
\begin{minipage}[]{.5\textwidth}
  \centering  
    \begin{algorithm}[H]
    \caption{Inference of \cmgpaggr}
    \label{alg:cmgpaggr_alg}
    \begin{algorithmic}
       \State {\bfseries Input:} $\srcN$ multi-resolution datasets $\{(\mathbf{X}_\srcn, \mathbf{Y}_\srcn) \}^{\srcN}_{\srcn=1}$, initial parameters $\theta$, 
    
       \State $\hat{\theta} \leftarrow \argmax_{\theta} \sum^\srcN_{\srcn=1} \ell(\mathbf{Y}_\srcn |  \theta) $
       
        \State $\mathbf{H} \leftarrow \sum^\srcN_{\srcn=1} (\nabla \ell(\mathbf{Y}_\srcn | \hat{\theta}) (\nabla \ell(\mathbf{Y}_\srcn | \hat{\theta}))^T$
        
        \State $\mathbf{J} \leftarrow \nabla^2   \ell(\mathbf{Y} | \hat{\theta})$
        
        \State $\cweight \leftarrow \begin{cases}
            \frac{|\hat{\theta}|}{\mathrm{Tr}[\mathbf{H}(\hat{\theta})^{-1} \mathbf{J}(\hat{\theta})]} \\
            \frac{ \mathrm{Tr}[\mathbf{H}(\hat{\theta}) \mathbf{J}(\hat{\theta})^{-1} \mathbf{H}(\hat{\theta})]}{\mathrm{Tr}[\mathbf{H}(\hat{\theta})]}
        \end{cases}$
        
        \State $\theta_1 \leftarrow \argmin_{\theta} \left( \sum^\srcN_{\srcn=1} \cweight \E_q\left[\ell(\mathbf{Y}_\srcn |  \theta)\right] + \mathcal{KL} \right)$
    
    \end{algorithmic}
    \end{algorithm}
\end{minipage}%
\caption{\textbf{Left}: Graphical model of \cmgpaggr for $\srcN$ observation processes each with $|P_\srcn|$ tasks. This allows \emph{multi-resolution learning} between and across tasks. \textbf{Right}: Inference for \cmgpaggr.}
\label{fig:mr_aggr_model_graphical_model}
\end{figure}

\subsection{Composite Likelihood Weights}
Under a misspecified model the asymptotic distribution of the MLE estimate converges to $\normal(\theta_0, \frac{1}{n} \mathbf{H}(\theta_0) \mathbf{J}(\theta_0)^{-1} \mathbf{H}(\theta_0) )$ where $\theta_0$ are the true parameters and $\mathbf{H}(\theta_0) = \frac{1}{n} \sum^N_{n=1} \nabla \ell(\mathbf{Y}|\theta_0) \nabla \ell(\mathbf{Y}|\theta_0)^T$, $\mathbf{J}(\theta_0) = \frac{1}{n} \sum^N_{n=1} \nabla^2 \ell(\mathbf{Y}|\theta_0)$ are the Hessian and Jacobian respectively. The form of the asymptotic variance is the \textit{sandwich information matrix} and it represents the loss of information in the MLE estimate due to the failure of Bartletts second identity \citep{Varin11anoverview}. 

Following \citet{2017arXiv170907616L} and \citet{Ribatet12bayesianinference} we write down the asymptotic posterior of \cmgpaggr as  $\normal(\theta_0, n^{-1} \cweight^{-1} \mathbf{H}(\theta_0))$. In practise we only consider a subset of parameters that present in all likelihood terms, such as the kernel parameters. Asymptotically one would expect the contribution of the prior to vanish causing the asymptotic posterior to match the limiting MLE. The composite weights $\cweight$ can be used to bring these distributions as close together as possible.  Approximating $\theta_0$ with the MLE estimate $\hat{\theta}$ and setting $\cweight^{-1} \mathbf{H}(\hat{\theta}) = \mathbf{H}(\hat{\theta}) \mathbf{J}(\hat{\theta})^{-1} \mathbf{H}(\hat{\theta})$ we can rearrange to find $\cweight$ and recover the magnitude correction of \citet{Ribatet12bayesianinference}. Instead if we take traces and then rearrange we recover the correction of \citet{2017arXiv170907616L}: 
\begin{equation} 
   \cweight_{\textrm{Ribatet}} = \frac{|\hat{\theta}|}{\mathrm{Tr}[\mathbf{H}(\hat{\theta})^{-1} \mathbf{J}(\hat{\theta})]} ~~~,~~~ \cweight_{\textrm{Lyddon}} = \frac{ \mathrm{Tr}[\mathbf{H}(\hat{\theta}) \mathbf{J}(\hat{\theta})^{-1} \mathbf{H}(\hat{\theta})]}{\mathrm{Tr}[\mathbf{H}(\hat{\theta})]}.
\end{equation}
\subsection{Inference}

In this section we a present a closed form variational lower bound for \cmgpaggr, the full details can be found in the Appendix. For computational efficiency we introduce inducing points (see \citep{pmlr-v5-titsias09a,Hensman:2013:GPB:3023638.3023667}) $\mathbf{U} = \{ \mathbf{u_q} \}^{Q}_{q=1}$ and $\mathbf{V} = \{ \mathbf{v_{p,q}} \}^{{P, Q}}_{p, q=1}$, for the latent GPs $\mathbf{f}$ and $\mathbf{W}$ respectively, where $\mathbf{u_q} \in \real^{M}$ and $\mathbf{v_{p,q}} \in \real^{M}$. The inducing points are at the corresponding locations $\mathbf{Z^{(u)}} = \{ \mathbf{Z^{(u)}_q} \}_{q=1}^Q, \mathbf{Z^{(v)}} = \{ \mathbf{Z^{(v)}_{p, q}} \}_{p,q=1}^{P,Q} ~~ \text{for} ~~ \mathbf{Z^{(\cdot)}_\cdot} \in \real^{M, D}$. We construct the augmented posterior and use the approximate posterior $q(\mathbf{u}, \mathbf{v}, \mathbf{f}, \mathbf{W}) = p(\mathbf{f}, \mathbf{W} | \mathbf{u}, \mathbf{v}) q(\mathbf{u}, \mathbf{v})$ where 
\begin{equation}
	\begin{aligned}
		q(\mathbf{u}, \mathbf{v}) =\sum^K_{k=1} \pi_k \prod^Q_{j=1} \normal(\mathbf{m}^{(\mathbf{u})}_j, \mathbf{S}^{(\mathbf{u})}_j) \cdot \prod^{P, Q}_{i, j=1} \normal(\mathbf{m}^{(\mathbf{v})}_{i, j}, \mathbf{S}^{(\mathbf{v})}_{i, j})
	\end{aligned}
\end{equation}
is a free form mixture of Gaussians with $K$ components. We follow the variational derivation of \cite{NIPS2014_5374, autogp_bonilla} and derive our expected log-likelihood $\text{ELL} = 
\sum_{\srcn=1}^{\srcN}\sum_{p=1}^P \sum_{n=1}^{N_a} \sum_{k=1}^K \text{ELL}_{a, p, n, k}$,
\begin{equation}
    \begin{aligned}
    &\text{ELL}_{a, p, n, k}
    = \pi_k \log \mathcal{N}\left(Y_{a, p, n}  \mid 
 \frac{1}{|\mathcal{S}_{a,n}|} \sum_{\mathbf{x} \in \mathcal{S}_{a,n}} \sum^Q_{q=1} \boldsymbol{\mu}^{(w)}_{k,p,q}(\mathbf{x}) \boldsymbol{\mu}^{(f)}_{k,q}(\mathbf{x})
, \sigma_{a, p}^2
\right) \\
&\qquad - \frac{\pi_k}{2\sigma_{a, p}^2}
\frac{1}{|S_{a, n}|^2}
\sum_{q=1}^Q \sum_{\mathbf{x}_1, \mathbf{x}_2} 
\boldsymbol{\Sigma}^{(w)}_{k,p,q}\boldsymbol{\Sigma}^{(f)}_{k,q} + 
    \boldsymbol{\mu}^{(f)}_{k,q}(\mathbf{x}_1)\boldsymbol{\Sigma}^{(w)}_{k,p,q}\boldsymbol{\mu}^{(f)}_{k,q} (\mathbf{x}_2) 
	\boldsymbol{\mu}^{(w)}_{k,p,q}(\mathbf{x}_1)\boldsymbol{\Sigma}^{(f)}_{k,q}\boldsymbol{\mu}^{(w)}_{k,p,q}(\mathbf{x}_2)
    \end{aligned}
\end{equation}

where $\boldsymbol{\Sigma}^{(\cdot)}_{\cdot,\cdot,\cdot}$ is evaluated at the points $\mathbf{x}_1$, $\mathbf{x}_2$.  and $\boldsymbol{\mu}^{(f)}_{k}$, $\boldsymbol{\mu}^{(w)}_{k,p}$, $\boldsymbol{\Sigma}^{(f)}_{k}$, $\boldsymbol{\Sigma}^{(w)}_{k,p}$ are respectively the mean and variance of $q_k(\mathbf{W}_{p})$, $q_k(\mathbf{f})$. To infer the composite weights we follow \citep{2017arXiv170907616L, Ribatet12bayesianinference} and first obtain the MLE estimate of $\theta$ by maximizing the likelihood in Eq. \ref{eq:main_aggr_likelihood}. The weights can then be calculated and the variational lowerbound optimised as in Alg. \ref{alg:cmgpaggr_alg} with $\mathcal{O}(\epochs\cdot (PQ+Q)NM^2)$ for $\epochs$ optimization steps until convergence. Our closed form ELBO generalizes prior state of the art of the GPRN (\citep{autogp_bonilla, pmlr-v31-nguyen13b, adams-stegle-2008a}) by extending to support multi-resolution data and allowing a free form mixture of Gaussians variational posterior. In the Appendix we also provide variational lower bounds for the positively-restricted GPRN form $\mathbf{Y}_p = \sum^{Q}_{q=1 }\exp(\mathbf{W}_{p,q}) \odot \mathbf{f}_q + \epsilon$ that we find can improve identifiability and predictive performance.


\subsection{Prediction}
Although the full predictive distribution of a specific observation process is not available in closed form, using the variational posterior we derive the predictive mean and variance, avoiding Monte Carlo estimates. The mean is simply $\E [\mathbf{Y}_{\srcn, p}^*] = \sum^K_k \pi_k E_k \left[ \mathbf{W}^*_p \right] \E_k[\mathbf{\hat{f}}^*]$, where $K$ is the number of components in the mixture of Gaussians variational posterior and $\pi_k$ is the $k$'th weight. We provide the predictive variance and full derivations in the appendix .

\begin{figure}
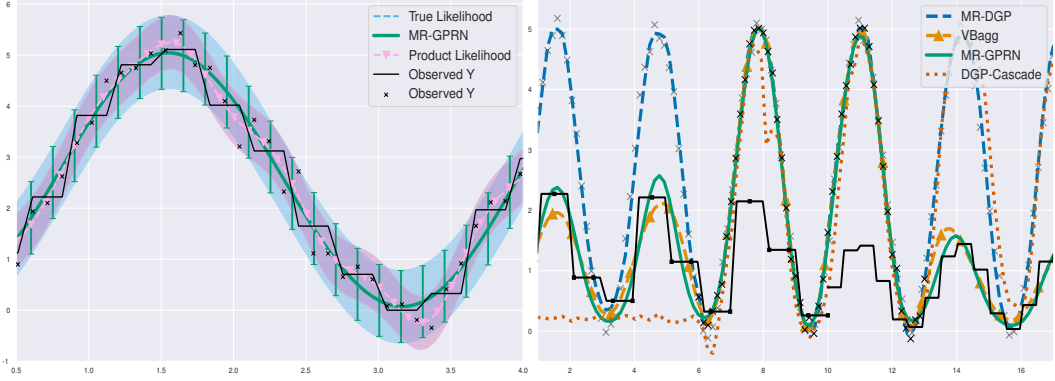

\centering
\begin{subfigure}[b]{0.5\linewidth}
  \centering
  \resizebox{1.0\linewidth}{5cm}{\input{appendix_images/compare_corrected.pgf}}
  \label{fig:compare_synth_correction}
\end{subfigure}%
\begin{subfigure}[b]{0.5\linewidth}
   \centering
    \resizebox{1.0\linewidth}{5cm}{\input{appendix_images/compare_baseline.pgf}}
   \label{fig:synth_compare_y1}
\end{subfigure}
\caption{\textbf{Left}: \cmgpaggr recovers the true predictive variance whereas assuming a product likelihood assumption leads to posterior contraction. \textbf{Right}: \cmgpdgp recovers the true predictive mean under a multi-resolution setting with scaling biases. Both \vbagg and \cmgpaggr fail as they propagate the bias. Black crosses and lines denote observed values. Grey crosses denote observations removed for testing.}
\label{fig:mcmc_posterior_contraction}
\end{figure}

\section{Multi-Resolution Deep Gaussian Processes (\cmgpdgp)}
\label{sec:mr_deep_gp}

\begin{figure}
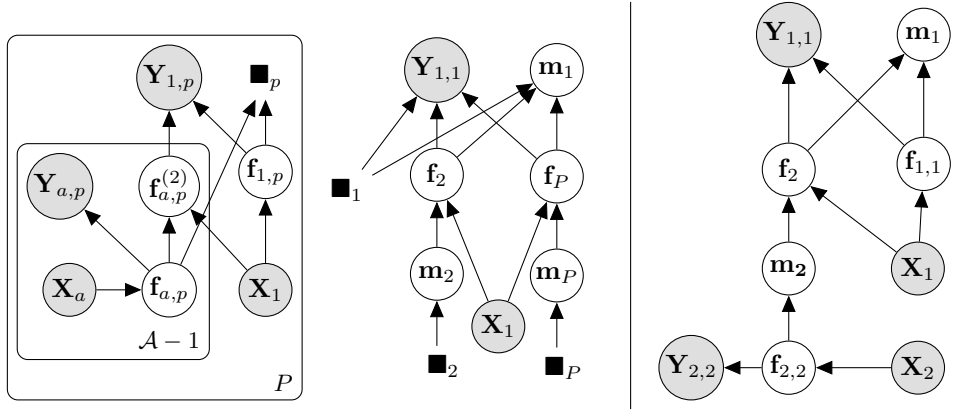

\centering
   \begin{tabular}{c|c}
\begin{subfigure}[t]{0.6\linewidth}
  \centering
    \tikz{
    
	\node[latent] (m1) {$\mathbf{m}_{1}$}; %
	\node[obs, left=of m1, xshift=0.2cm] (y1) {$\mathbf{Y}_{1, 1}$}; %

	\node[latent, below=of y1, yshift=0.4cm] (f2) {$\mathbf{f}_{2}$}; %
	\node[latent, below=of f2, yshift=0.4cm] (m2) {$\mathbf{m}_{2}$}; %
	
	\node[obs, right=of m2, xshift=-0.9cm, yshift=-0.7cm] (x1main) {$\mathbf{X}_1$}; %
	
	\node[below=of m2, yshift=0.4cm] (m2factor1) {$\phantom{o}\blacksquare_2$}; %

	\node[latent, below=of m1,yshift=0.3cm] (fp) {$\mathbf{f}_{P}$}; %
	\node[latent, below=of fp, yshift=0.4cm] (mp) {$\mathbf{m}_{P}$}; %
	\node[below=of mp, yshift=0.4cm] (mpfactor1) {$\phantom{o}\blacksquare_{P}$}; %
	
	\node[left=of f2,xshift=0.5cm, yshift=-0.2cm] (factor1) {$\blacksquare_1$}; %
	
	\node[left=of y1, yshift=-0.1cm, xshift=-0.5cm] (factorpp) {$\blacksquare_p$}; %
	\node[obs, left=of factorpp, xshift=0.5cm] (yp) {$\mathbf{Y}_{1, p}$}; %
	\node[latent, below=of factorpp,yshift=0.4cm] (f1p) {$\mathbf{f}_{1, p}$}; %
	\node[obs, below=of f1p, yshift=0.15cm] (xp) {$\mathbf{X}_1$}; %
	
	\node[latent, below=of yp,yshift=0.4cm] (fap1) {$\mathbf{f}^{(2)}_{\srcn, p}$}; %
	\node[latent, below=of fap1, yshift=0.4cm] (fap2) {$\mathbf{f}_{\srcn, p}$}; %

	\node[obs, left=of fap1, xshift=0.4cm] (yhp) {$\mathbf{Y}_{\srcn, p}$}; %
	\node[obs, left=of fap2, xshift=0.4cm] (xhp) {$\mathbf{X}_\srcn$}; %
	
	\plate[] {pap} {(fap1)(fap2)(yhp)(xhp)} {$\srcN-1$};%
	
	\plate[] {pmp} {(f1p)(fap1)(fap2)(yhp)(xp)(yp)(xhp)(factorpp)(pap)} {$P$};%
	
	\edge {fap2} {fap1}; %
	\edge {fap2} {yhp}; %
	\edge {fap1} {yp}; %
	\edge {xhp} {fap2}; %
	\edge {fap2} {factorpp}; %
	\edge {f1p} {factorpp}; %
	\edge {f1p} {yp}; %
	\edge {xp}{f1p}; %
	\edge {xp}{fap1}; %
	\edge {x1main} {f2}; %
	\edge {x1main} {fp}; %

	\edge {f2} {m1} ; %
	\edge {f2} {y1} ; %
	\edge {fp} {m1} ; %
	\edge {fp} {y1} ; %
	\edge {m2} {f2} ; %
	\edge {mp} {fp} ; %
	\edge {m2factor1} {m2} ; %
	\edge {mpfactor1} {mp} ; %
	\edge {factor1} {m1} ; %
	\edge {factor1} {y1} ; %
}
\end{subfigure}&
\begin{subfigure}[b]{0.3\linewidth}
   \centering
    \tikz{
        \node[latent] (g2) {$\mathbf{m}_1$}; %
        \node[obs, left=of g2] (y2) {$\mathbf{Y}_{1, 1}$}; %

        \node[latent, below =of g2] (f12) {$\mathbf{f}_{1, 1}$}; %
        \node[latent, below =of y2] (f2) {$\mathbf{f}_{2}$}; %

        \node[latent,below=of f2, yshift=0.4cm] (g1) {$\mathbf{m_2}$}; %
        \node[latent, below=of g1, yshift=0.4cm] (f11) {$\mathbf{f}_{2, 2}$}; %
        \node[obs, right=of f11] (x1) {$\mathbf{X}_2$}; %
        \node[obs, left=of f11, xshift=0.5cm] (y1) {$\mathbf{Y}_{2,2}$}; %
        
        \node[obs, right=of g1] (x2) {$\mathbf{X}_1$}; %

        \edge {x1} {f11}; %
        \edge {f11} {y1} ; %
        \edge {f11} {g1} ; %
        \edge {g1} {f2} ; %
        \edge {x2} {f2} ; %
        \edge {x2} {f12} ; %
        \edge {f2} {g2} ; %
        \edge {f12} {g2} ; %
        \edge {f2} {y2} ; %
        \edge {f12} {y2} ; %
}
\end{subfigure}%
  \end{tabular}
\caption{\textbf{Left}: General plate diagram of \cmgpdgp for $\srcN$ observation processes across $P$ tasks with noise variances omitted. For notational simplicity we have assumed that the target resolution is $a=1$ and we use $\blacksquare_p$ to depict each of the sub-plate diagrams defined on the LHS. \textbf{Right}: A specific instantiation of an \cmgpdgp for 2 tasks and 2 observation processes (resolutions) with a target process $\mathbf{Y}_{1,1}$ as in the \emph{inter}-task multi-resolution PM10, PM25 experiment in Section 4.}
\label{fig:mr_dgp_plate}
\end{figure}

We now introduce \cmgpdgp, a deep instantiation of the framework  which extends the deep GP (\dgp) model of \citet{pmlr-v31-damianou13a} into a tree-structured multi-resolution construction, Fig. \ref{fig:mr_dgp_plate}. For notational convenience henceforth we assume that $p=1$ is the target task and that $a=1$ is the highest resolution and the one of primary interest. We note that this need not be the case and the relevant expressions can be trivially updated accordingly.

\subsection{Model Specification}

First we focus on the case when $P=1$ and then generalize to an arbitrary number of tasks. We place $\srcN$ independent ``Base" GPs $\{ \mathbf{f_{a,p}} \}^{\srcN}_{a=1}$ on each of the $\srcN$ datasets within task $p$ that model their corresponding resolution independently.  Taking $\srcn=1$ to be the target observation process we now construct $\srcN-1$ DGPs that map from these base GPs $\{ \mathbf{f_{a,p}} \}^{\srcN}_{a=2}$ to the target process $a=1$ while learning an input-dependent mapping between observation processes. These DGPs are local experts that capture the information contained in each resolution for the target observation process. Every GP has an explicit likelihood which enables us to estimate and predict at every resolution and task while allowing for biases between observation processes to be corrected, see Fig. \ref{fig:mcmc_posterior_contraction}.

More formally, the likelihood of the \cmgpdgp with one task is $p(\mathbf{Y}_p|\mathbf{F}_p)$=
\begin{equation}
	\underbrace{\prod^\srcN_{\srcn=2} \normal(\mathbf{Y}_{1,p} | \frac{1}{|S_a|}\int_{S_a}\mathbf{f}^{(2)}_{a,p} (\mathbf{x}) \mathop{d \mathbf{x}}, \sigma^2_{\srcn,p}) p(\mathbf{f}^{(2)}_{a,p}|\mathbf{f}_{a,p})  }_{\text{Deep GPs}} \cdot \underbrace{\prod^\srcN_{\srcn=1} \normal((\mathbf{Y}_{a,p} | \frac{1}{|S_a|}\int_{S_a}\mathbf{f}_{a,p} (\mathbf{x}) \mathop{d \mathbf{x}}, \sigma^2_{\srcn,p})  p(\mathbf{f}_{a,p})}_{\text{Base GPs}}
\end{equation}
where $\mathbf{f}_{a,p} \sim \mathcal{GP}(0, \mathbf{K}_{a,p})$ and we have stacked all the observations and latent GPs into $\mathbf{Y}_p$ and $\mathbf{F}_p$ respectively. Each of the likelihood components is a special case of the multi-resolution likelihood in \eq \ref{eq:main_aggr_likelihood} (where $Q=1$ and the latent GPs $\mathbf{W}$ are constant) and we discretize the integral in the same fashion. Similarly to the deep multi-fidelity model of \citep{2019arXiv190307320C} we define each DGP as: 
\begin{equation}
	p(\mathbf{f}^{(2)}_{a,p}|\mathbf{f}_{a,p}) = \normal(0, \mathbf{K}^{(2)}_{a,p}((\mathbf{f}_{a,p}, \mathbf{X}_1), (\mathbf{f}_{a,p}, \mathbf{X}_1))) 
\end{equation}
where $\mathbf{X}_1$ are the covariates of the resolution of interest in our running example and allow each \dgp to learn a mapping, between any observation process $a$ and the target one, that varies across $\mathbf{X}_1$. We now have $\srcN$ independent DGPs modelling $\mathbf{Y}_{1,p}$ with separable spatio-temporal kernels at each layer. The observation processes are not only at varying resolutions, but could also be partially overlapping or disjoint. This motivates treating each GP as a local model in a mixture of GP experts \cite{NIPS2008_3395}. Mixture of GP experts typically combine the local GPs in two ways: either through a gating network \cite{NIPS2001_2055} or through weighing the local GPs \cite{Deisenroth:2015:DGP:3045118.3045276,pmlr-v32-nguyena14}. We employ the mixing weight approach in order to avoid the computational burden of learning the gating work. We define the mixture $\mathbf{m}_p = \beta_1 \odot \mathbf{f}_{1, p} + \sum^{\srcN}_{a=1} \beta_a \odot \mathbf{f}^{(2)}_{a, p}$ where the weight captures the reliability of the local GPs (or is set to 1 if the mixture is a singleton). The reliability is defined by the resolution and support of the base GPs and is naturally achieved by utilising the normalised log variances of the base GPs as $\beta_a = (1-\mathbf{V}_a) \sum^a_i \mathbf{V}_i$. We provide the full justification and derivation for these weights in the appendix.

We can now generalize to an arbitrary number of tasks. For each task we construct a mixture of experts $\mathbf{m}_p$ as described above. For tasks $p>1$ we learn the mapping from $\mathbf{m}_p$ to the target observation process $\mathbf{Y}_{1,1}$. This defines another set of local GP experts that is combined into a mixture with DGP experts. In our experiments we set $\mathbf{m}_p$ for $p>1$ to be a simple average and for $\mathbf{m}_1$ we use our variance derived weights. This formulation naturally handles biases between the mean of different observations processes and each layer of the DGPs has a meaningful interpretation as it is modelling a specific observation process.

\subsection{Augmented Posterior}

Due to the non-linear forms of the parent GPs within the DGPs marginalising out the parent GPs is generally analytically intractable. Following \cite{NIPS2017_7045} we introduce inducing points $\mathbf{U} =  \{ \mathbf{u}_p \}^{P}_{p=2} \cup \{ \mathbf{u}^{(2)}_{a,p},\mathbf{u}_{a,p} \}^{P, \srcN}_{a,p=1}$ where each $\mathbf{u}^{(\cdot)}_{\cdot, \cdot} \in \real^{M}$ and inducing locations $\mathbf{Z} =  \{ \mathbf{Z}_p \}^{P}_{p=2} \cup \{ \mathbf{Z}^{(2)}_{a,p},\mathbf{Z}_{a,p} \}^{P, \srcN}_{a,p=1}$ where $\mathbf{Z}_p, \mathbf{Z}^{(2)}_{a,p} \in \real^{M \times (D+1)}$ and $\mathbf{Z}_{a,p} \in \real^{M \times D}$. The augmented posterior is now simply $p(\mathbf{Y}, \mathbf{F}, \mathbf{M}, \mathbf{U}) = p(\mathbf{Y}|\mathbf{F})p(\mathbf{F}, \mathbf{M} | \mathbf{U}) p(\mathbf{U})$ (with slight notation abuse) where each $p(\mathbf{u}^{(\cdot)}_{\cdot, \cdot}) = \normal(0, \mathbf{K}^{(\cdot)}_{\cdot, \cdot})$. Full details are provided in the Appendix.

\subsection{Inference}

Following \cite{NIPS2017_7045} we construct an approximate augmented posterior that maintains the dependency structure between layers:
\begin{equation}
\begin{aligned}
	q(\mathbf{M}, \mathbf{F}, \mathbf{U}) &= p(\mathbf{M}, \mathbf{F} | \mathbf{U}) \prod^{P}_{p=2} q(\mathbf{u}_p) \cdot \prod^{P}_{p=1} \prod^\srcN_{\srcn = 1} q(\mathbf{u}^{(2)}_{a,p}) q(\mathbf{u}_{a,p}) &\\ 
\end{aligned}
\end{equation}
where each $q(\mathbf{u}^{(\cdot)}_{\cdot,\cdot})$ are independent free-form Gaussian $\normal(\mathbf{m}^{(\cdot)}_{\cdot,\cdot}, \mathbf{S}^{(\cdot)}_{\cdot,\cdot})$ and the conditional is
\begin{equation}
	p(\mathbf{F}, \mathbf{M} | \mathbf{U})= \prod^P_{p=2} p(\mathbf{f}_p|\mathbf{m}_p, \mathbf{u}_p) p(\mathbf{m}_p | \uParent(\mathbf{m}_p)) \cdot \prod^P_{p=1} p(\mathbf{f}_{1,p} | \mathbf{u}_{1,p}) \prod^\srcN_{a=2}  
	p(\mathbf{f}^{(2)}_{a,p} | \mathbf{f}_{a,p}, \mathbf{u}^{(2)}_{a,p})  p(\mathbf{f}_{a,p}| \mathbf{u}_{a,p}).
\end{equation}
We use $\uParent(\cdot)$ to denote the set of parent GPs of a given GP and $ \numL(\mathbf{f})$ to denote the depth of DGP $\mathbf{f}$, $p(\mathbf{m}_p | \uParent(\mathbf{m}_p)) = \normal(\sum^\srcN_\srcn \mathbf{w}_{a,p}\mu_{a,p}, \sum^\srcN_\srcn \mathbf{w}_{a,p} \Sigma_{a,p} \mathbf{w}_{a,p})$ and $\mu_{a,p}, \Sigma_{a,p}$ are the mean and variance of the relevant DGPs. Note that the mixture $\mathbf{m}_1$ combines all the DGPs at the top layer of the tree-hierarchy and hence it only appears in the predictive distribution of \cmgpdgp. All other terms are standard sparse GP conditionals and are provided in the Appendix. The ELBO is be simply derived as 
\begin{equation}
   \mathcal{L}_{\cmgpdgp} = \underbrace{\expected_{q(\mathbf{M}, \mathbf{F}, \mathbf{U})} \left[ \log p(\mathbf{Y} | \mathbf{F}) \right]}_{\text{ELL}} + \underbrace{\expected_{q(\mathbf{U})} \left[ \log \frac{P(\mathbf{U})}{q(\mathbf{U})} \right]}_{\text{KL}}
\end{equation}
where the KL term is decomposed into a sum over all inducing variables $\mathbf{u}^{(\cdot)}_{\cdot, \cdot}$. The expected log likelihood (ELL) term decomposed across all $\mathbf{Y}$:
\begin{equation}
	\sum^P_{p=2} \expected_{q(\mathbf{f}_p)} \left[ \log p(\mathbf{Y}_{1,1} | \mathbf{f}_p) \right]+ \sum^P_{p=1} \sum^\srcN_\srcn \left[ \expected_{q(\mathbf{f}^{(2)}_{a, 1})} \left[\log p(\mathbf{Y}_{1,p} | \mathbf{f}^{(2)}_{a, 1})\right] + \expected_{q(\mathbf{f}_{a,p})} \left[\log p(\mathbf{Y}_{a,p} | \mathbf{f}_{a,p})\right] \right].
\end{equation}
For each ELL component the marginal $q(\mathbf{f}^{(\cdot)}_{\cdot, \cdot})$ is required. Because the base GPs are Gaussian, sampling is straightforward and the samples can be propagated through the layers, allowing the marginalization integral to be approximated by Monte Carlo samples. We use the reparametization trick to draw samples from the variational posteriors \citep{kingma_auto_encoder}. The inference procedure is given in Alg. \ref{alg:cmgp_dgp}.

\begin{algorithm}[tb]
   \caption{Inference procedure for \cmgpdgp}
   \label{alg:vcmgpdgp_alg}
\begin{algorithmic}
   \State {\bfseries Input:} $P$ multi-resolution datasets $\{(\mathbf{X}_p, \mathbf{Y}_p) \}^{P}_{p=1}$, initial parameters $\theta_0$, 

\Procedure{Marginal}{$\mathbf{f}$,$\mathbf{X}$, l, L}
   \If{$l = L$}
      \State \textbf{return} $ q(\mathbf{f}|\mathbf{X})$
   \EndIf
   \State  $ q(\parent(\mathbf{f})|\mathbf{X}) \gets $ \acro{Marginal}($\parent(\mathbf{f})$, $\mathbf{X}$, $l+1$, $\numL(\parent(\mathbf{f})))$
   \State \textbf{return} $\frac{1}{S} \sum^S_{s=1} p(\mathbf{f} | \mathbf{f}^{(s)},\mathbf{X})) $ where $\mathbf{f}^{(s)} \sim  q(\parent(\mathbf{f})|\mathbf{X})$ 
\EndProcedure%

\State $\theta_1 \leftarrow \argmin\limits_{\theta} \left[  \E_{\{\acro{Marginal}(\mathbf{f}_p, \mathbf{X}_\srcn, 0, \numL(\mathbf{f}_p)) \}^P_{p=1} } \left[\log p(\mathbf{Y} | \mathbf{F}, \mathbf{X}, \theta)\right] + \mathcal{KL}(q(\mathbf{U}) || p(\mathbf{U})) \right]$ 

\end{algorithmic}
\label{alg:cmgp_dgp}
\end{algorithm}

\subsection{Prediction}

\textbf{Predictive Density}. To predict at $\mathbf{x^*} \in \real^D$ in the target resolution $a=1$ we simply approximate the predictive density $q(\mathbf{\mixf}^*_1)$ by sampling from the variational posteriors and propagating the samples $\mathbf{f}^{(s)}$ through all the layers of our \cmgpdgp structure:
\begin{equation}
    q(\mathbf{\mixf}^*_1) = \int q(\mathbf{\mixf}^*_1 | \uParent(\mathbf{\mixf}^*_1)) \prod_{\mathbf{f} \in \uParent(\mathbf{\mixf}^*_1)} q(\mathbf{f}) \mathop{d\uParent(\mathbf{\mixf}^*_1)} \approx \frac{1}{S} \sum^S_{s=1} q(\mathbf{\mixf}^*_1 | \{ \mathbf{f}^{(s)} \}_{\mathbf{f} \in \uParent(\mathbf{\mixf}^*_1)})
\end{equation}
In fact while propagating the samples through the tree structure the model naturally predicts at every resolution $a$ and task $p$ for the corresponding input location.

\section{Related Work}
\label{sec:lit}

Gaussian processes (GPs) are the workhorse for spatio-temporal modelling in spatial statistics \citep{gelfand2010handbook} and in machine learning \citep{gpml} with the direct link between multi-task GPs and Linear Models of Coregionalisation (LCM) reviewed by \citet{alvarez2012kernels}. Heteroscedastic GPs \citep{Lazaro-Gredilla:2011:VHG:3104482.3104588} and recently proposed deeper compositions of GPs for the multi-fidelity setting \citep{Perdikaris_Nonlinear_multi_fidelity,2019arXiv190307320C,perdikaris2015multi} assume that all observations are of the same resolution. In spatial statistics the related \emph{change of support} problem has been approached through Markov Chain Monte Carlo approximations and domain discretizations  \citep{fuentes2005model,gelfand2010handbook}. A recent exception to this is the work by \citet{2018arXiv180902010S} that solves the integral for squared exponential kernels but only considers observations from one resolution and cannot handle additional input features. Independently and concurrently, \cite{yousefi2019multitask} have recently proposed a multi-resolution LCM model that is similar to our \cmgpaggr model without dependent observation processes and composite likelihood corrections but instead a focus on improved estimation of the area integral and non-Gaussian likelihoods. Finally, we note that the multiresolution GP work by \citet{Fox:2012:MGP:2999134.2999217} defines a \dgp construction for non-stationary models that is more akin to multi-scale modelling \citep{multiscaleGP}. This line of research typically focuses on learning multiple kernel lengthscales to explain both broad and fine variations in the underlying process and hence  cannot handle multi-resolution observations .


\section{Experiments}
\label{sec:exp}

\begin{figure}
\centering
\begin{subfigure}[t]{0.32\linewidth}
   \centering
    \caption*{\cmgpdgp}
   \includegraphics[width=\textwidth,height=\textwidth]{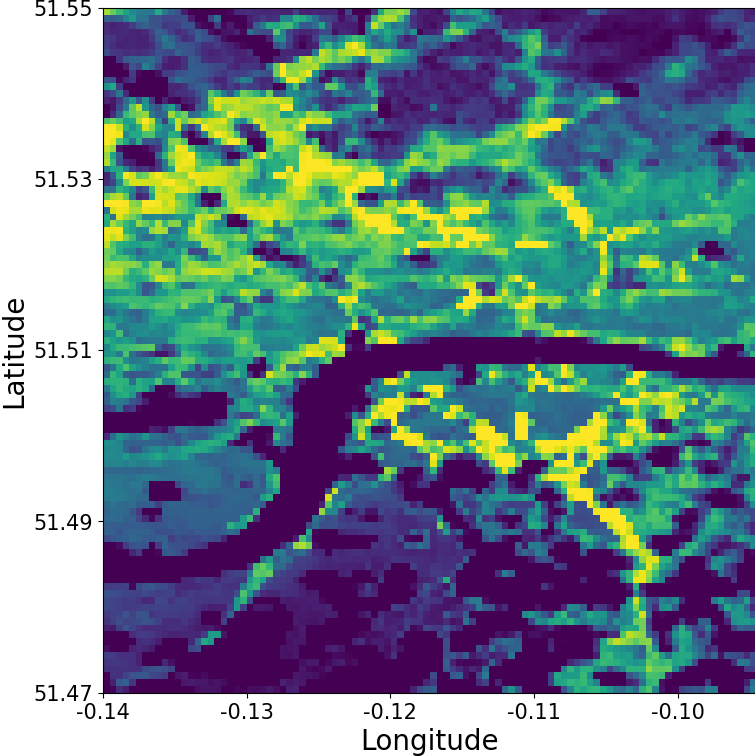}
   \label{fig:synth_compare_y1}
\end{subfigure}%
~
\begin{subfigure}[t]{0.32\linewidth}
   \centering
   \caption*{\vbagg}
   \includegraphics[width=0.9\textwidth,height=\textwidth]{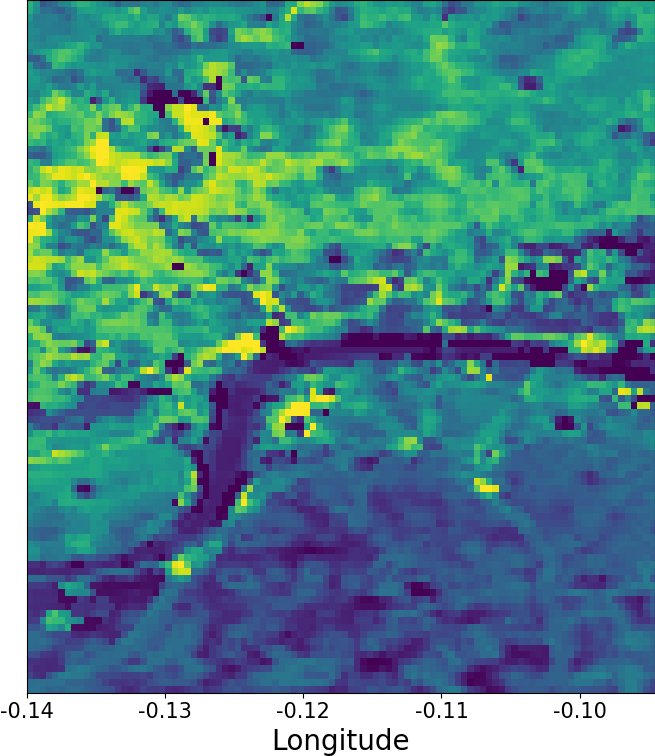}
   \label{fig:synth_compare_y1}
\end{subfigure}%
~ 
\begin{subfigure}[t]{0.32\linewidth}
   \centering
   \caption*{\centerpoint}
   \includegraphics[width=\textwidth,height=\textwidth]{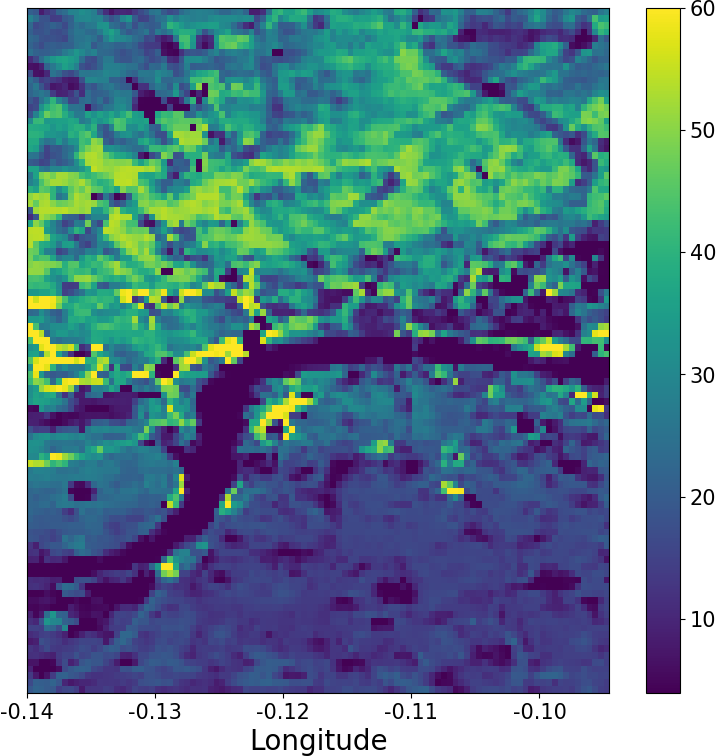}
   \label{fig:synth_compare_y1}
\end{subfigure}%

\begin{subfigure}[t]{0.32\linewidth}
   \centering
   \includegraphics[width=\textwidth,height=\textwidth]{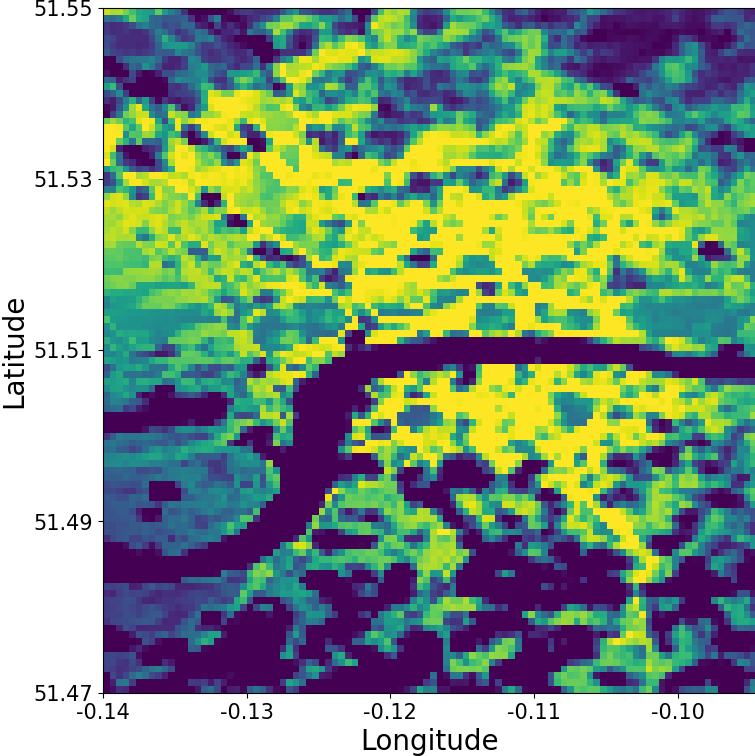}
   \label{fig:synth_compare_y1}
\end{subfigure}%
~
\begin{subfigure}[t]{0.32\linewidth}
   \centering
   \includegraphics[width=0.9\textwidth,height=\textwidth]{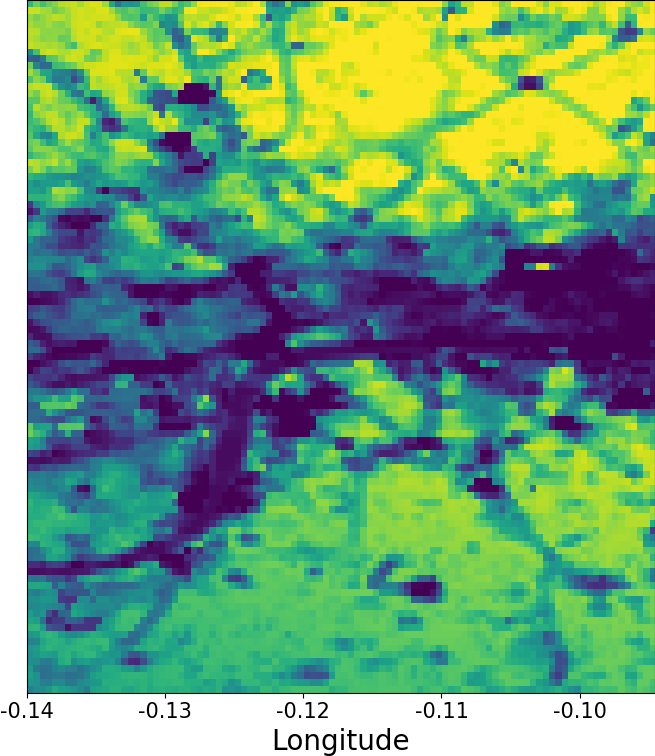}
   \label{fig:synth_compare_y1}
\end{subfigure}%
~
\begin{subfigure}[t]{0.32\linewidth}
   \centering
   \includegraphics[width=\textwidth,height=\textwidth]{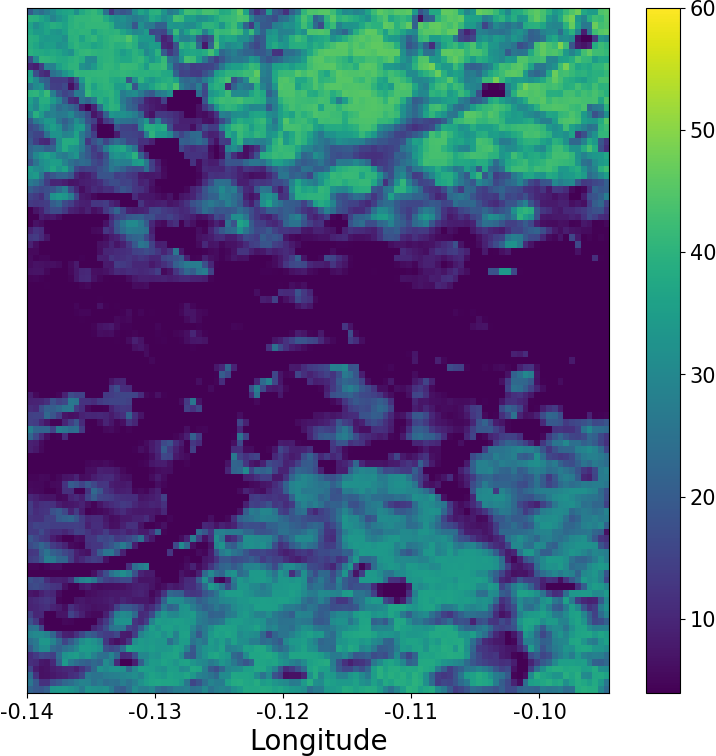}
   \label{fig:synth_compare_y1}
\end{subfigure}%

\caption{Spatio-temporal estimation and forecasting of NO$_2$ levels in London. \textbf{Top Row}: Spatial slices from \cmgpaggr, \vbagg and \centerpoint respectively at 19/02/2019 11:00:00 using observations from both LAQN and the satellite model (low spatial resolution). \textbf{Bottom Row}: Spatial slices at the base resolution from the same models at 19/02/2019 17:00:00 where \emph{only} observations from the satellite model are present. }
\label{fig:sat_no2_st}
\end{figure}


We demonstrate and evaluate the MRGPs on synthetic experiments and the challenging problem of estimating and forecasting air pollution in the city of London. We compare against \vbagg \cite{law2018variational} and two additional baselines. The first, \centerpoint, is a GPRN modified to support multi-resolution data by taking the center point of each aggregation region as the input. The second, \dgpcascade is a \cmgpdgp but instead of a tree structured \dgp as in \uFig \ref{fig:mr_dgp_plate} we construct a cascade to illustrate the benefits of the tree composition and the mixture of experts approach of \cmgpdgp. Experiments are coded\footnote{Codebase and datasets to reproduce results are available at \url{https://github.com/ohamelijnck/multi_res_gps}} in \textit{TensorFlow} and we provide additional analysis in the Appendix.

\textbf{Dependent observation processes}: We provide additional details of the dependent observation processes experiment in the left of Fig. \ref{fig:mcmc_posterior_contraction} in the Appendix.

\textbf{Biased observation processes:}. To demonstrate the ability of \cmgpdgp in handling  biases across observation processes we construct 3 datasets from the function $\mathbf{y}=s \cdot 5 \sin(\mathbf{x})^2+0.1\epsilon$ where $\epsilon \sim\normal(0, 1)$. The first $\mathbf{X}_1, \mathbf{Y}_1$ is at resolution $\res_1= 1$ in the range x=[7,12] with a scale $s=1$. The second is at resolution of $\res_2=5$ between x=[-10, 10] with a scale $s=0.5$ and lastly the third is at resolution of $\res_3=5$ x=[10, 20] with a scale $s=0.3$. The aim is to predict $\mathbf{y}$ across the range [-10, 20] and the results are shown in \uTable \ref{table:no2_sythn} and \uFig \ref{fig:mcmc_posterior_contraction}. $\cmgpdgp$ significantly outperforms all of the four alternative approaches as it is learning a forward \emph{mapping} between observation processes, e.g. $\mathbf{f}_2^{(2)}$ in Fig. \ref{fig:mr_dgp_plate}, and is not just trusting and propagating the mean.

\textbf{Training}. When training both \cmgpaggr and \vbagg we first jointly optimize the variational and hyper parameters while keeping the likelihood variances fixed and then jointly optimize all parameters together. For \cmgpdgp we first optimize layer by layer and then jointly optimize all parameters together, see Appendix. We find that this helps to avoid early local optima. 

\textbf{\emph{Inter}-task multi-resolution: modelling of PM$_{10}$ and PM$_{25}$ in London}: In this experiment we consider multiple tasks with different resolutions. We jointly model PM$_{10}$ and PM$_{25}$ at a specific LAQN location in London. The site we consider is \emph{RB7} in the date range 18/06/2018 to 28/06/2018. At this location we have hourly data from both PM$_{10}$ and PM$_{25}$. To simulate having multiple resolutions we construct 2, 5, 10 and 24 hour aggregations of PM$_{10}$ and remove a 2 day region of PM$_{25}$ which is the test region.  The results from all of our models in \uTable \ref{table:mt_pm15} demonstrate the ability to successfully learn the multi-task dependencies. Note that \centerpoint fails, e.g. \uTable \ref{table:no2_sythn}, when the sampling area cannot be approximated by a single center point due the scale of the underlying process. 

\begin{table}
   \caption{\emph{Inter}-task multi-resolution. Missing data predictive MSE on PM$_{25}$ from \cmgpaggr, \cmgpdgp and baseline \centerpoint for 4 different aggregation levels of PM$_{10}$. \vbagg is inapplicable in this experiment as it is a single-task approach.}
   \label{table:mt_pm15}
   \centering
   \begin{filecontents*}{tables/multi_res_ys.csv}
Model,2,5,10,24
\centerpoint,4.67 $\pm$ 0.74,5.04 $\pm$ 0.45,5.26 $\pm$ 0.91,5.72 $\pm$ 0.91
\cmgpaggr,4.54 $\pm$ 0.93,5.09 $\pm$ 1.04,4.96 $\pm$ 1.07,5.32 $\pm$ 1.14
\cmgpdgp,5.14 $\pm$ 1.28,4.81 $\pm$ 1.06,4.61 $\pm$ 1.43,5.42 $\pm$ 1.15
\end{filecontents*}
\pgfplotstabletypeset[
      col sep=comma,
      string type,
      every head row/.style={
        before row={
            \\ \toprule
            \multicolumn{1}{c}{\multirow{2}{*}{Model}} & \multicolumn{4}{c}{PM$_{10}$ Resolution} \\
            \cmidrule(lr){2-5}
        },
        after row={
            \hline
        },
      },
      every last row/.style={after row=\bottomrule},
      columns/Model/.style={column name=\space, column type={l}},
      columns/2/.style={column name=2 Hours, column type={l}},
      columns/5/.style={column name=5 Hours, column type={l}},
      columns/10/.style={column name=10 Hours, column type={l}},
      columns/24/.style={column name=24 Hours, column type={l}},
   ]{tables/multi_res_ys.csv}
\end{table}

\textbf{\emph{Intra}-task multi-resolution: spatio-temporal modelling of $\text{NO}_2$ in London}: In this experiment we consider the case of a single task but with multiple multi-resolution observation processes. First we use observations coming from ground point sensors from the London Air Quality Network (LAQN). These sensors provide hourly readings of $\text{NO}_2$. Secondly we use observations arising from a global satellite model \cite{sat_data} that provide hourly data at a spatial resolution of $7$km $\times$ $7$km and provide 48 hour forecasts.  We train on both the LAQN and satellite observations from 19/02/2018-20/02/2018 and the satellite ones from 20/02/2018-21/02/2018. We then predict at the resolution of the LAQN sensors in the latter date range. To calculate errors we predict for each LAQN sensor site, and find the average and standard deviation across all sites.

We find that \cmgpdgp is able to substantially outperform both \vbagg, \cmgpaggr and the baselines, \uTable \ref{table:no2_sythn} (left), as it is learning the forward mapping between the low resolution satellite observations and the high resolution LAQN sensors, while handling scaling biases. This is further highlighted in the bottom of \uFig \ref{fig:sat_no2_st} where \cmgpdgp is able to retain high resolution structure based only on satellite observations whereas \vbagg and \centerpoint over-smooth.

\begin{table}
    \caption{\emph{Intra}-task multi-resolution. \textbf{Left}: Predicting $\text{NO}_2$ across London (Fig. \ref{fig:sat_no2_st}). \textbf{Right}: Synthetic experiment results (Fig. \ref{fig:mcmc_posterior_contraction}) with three observations processes and scaling bias.}
    \begin{subtable}[t]{0.4\textwidth}
       \label{table:sat_no2}
       \centering
       \begin{filecontents*}{tables/no2_space_time_results.csv}
Model,RMSE,MAPE
Single GP,20.55 $\pm$ 9.44,0.8 $\pm$ 0.16
\centerpoint,18.74 $\pm$ 12.65,0.65 $\pm$ 0.21
\vbagg,16.16 $\pm$ 9.44,0.69 $\pm$ 0.37
\cmgpaggr w/o CL,12.97 $\pm$ 9.22,0.56 $\pm$ 0.32
\cmgpaggr w CL,11.92 $\pm$ 6.8,0.45 $\pm$ 0.17
\cmgpdgp,\textbf{6.27} $\pm$ \textbf{2.77},\textbf{0.38} $\pm$ \textbf{0.32}

\end{filecontents*}
\pgfplotstabletypeset[
          col sep=comma,
          string type,
          columns/model/.style={column name=Model, column type={l}},
          columns/rmse/.style={column name=RMSE, column type={l}},
          columns/mse/.style={column name=MSE, column type={l}},
          every head row/.style={before row=\\ \toprule,after row=\hline},
          every last row/.style={after row=\bottomrule}
       ]{tables/no2_space_time_results.csv}
    \end{subtable}
         \hspace{6em}
    \begin{subtable}[t]{0.40\textwidth}
    \label{table:synth_exp}
    \centering
    \begin{filecontents*}{tables/bias_exp.csv}
Model,RMSE,MAPE
\dgpcascade,2.12,0.16
\vbagg,1.68,0.14
\cmgpaggr,1.6,0.14
\cmgpdgp,\textbf{0.19},\textbf{0.02}
\end{filecontents*}
\pgfplotstabletypeset[
      col sep=comma,
      string type,
      columns/model/.style={column name=Model, column type={l}},
      columns/rmse/.style={column name=RMSE, column type={l}},
      columns/mse/.style={column name=MSE, column type={l}},
      every head row/.style={before row=\\ \toprule,after row=\hline},
      every last row/.style={after row=\bottomrule}
   ]{tables/bias_exp.csv}
    \end{subtable}
    \label{table:no2_sythn}
\end{table}

\section{Conclusion}
\label{sec:conc}

We offer a framework for evidence integration when observation processes can have varying \emph{inter-} and \emph{intra-task} sampling resolutions, dependencies, and different signal to noise ratios. Our motivation comes from a challenging and impactful problem of hyper-local air quality prediction in the city of London, while the underlying multi-resolution multi-sensor problem is general and pervasive across modern spatio-temporal settings and applications of machine learning. We proposed both shallow mixtures and deep learning models that generalise and outperform the prior art, correct for posterior contraction, and can handle biases in observation processes such as discrepancies in the mean. Further directions now open up to robustify the multi-resolution framework against outliers and against further model misspecification by exploiting ongoing advances in generalized variational inference  \citep{2019arXiv190402063K}. Finally an open challenge remains on developing continuous model constructions that avoid domain discretization, as in  \citep{pmlr-v71-adelsberg18a,yousefi2019multitask}, for more complex  settings.

\subsubsection*{Acknowledgements}

O. H., T. D and K.W. are funded by the Lloyd's Register Foundation programme on Data Centric Engineering through the London Air Quality project. This work is supported by The Alan Turing Institute for Data Science and AI under EPSRC grant EP/N510129/1 in collaboration with the Greater London Authority. We would like to thank the anonymous reviewers for their feedback and Libby Rogers, Patrick O'Hara and Daniel Tait for their help on multiple aspects of this work.

\bibliographystyle{apalike}
\bibliography{main}

\clearpage
\begin{appendix}

\section{\cmgpaggr}

In this section we provide the full derivation of the variational lower bound for  \cmgpaggr. Recall that we have a multi-resolution data set across $P$ tasks and we model each task as
\begin{equation}
	\mathbf{Y}_{a,p} = \frac{1}{|\mathcal{S}_a|} \int_{\mathcal{S}_a}\sum^Q_{q=1} \mathbf{W}_{p,q}(\mathbf{x})\mathbf{f}_{q}(\mathbf{x}) \mathop{d\mathbf{x}} + \epsilon_{a,p}
\end{equation}
where $\epsilon_{a,p} \sim \normal(0, \sigma^2_{a,p} \mathbf{I})$. The joint
density $p(\mathbf{Y}, \mathbf{W}, \mathbf{f})$ is then proportional to
\begin{equation}
	\underbrace{\prod^{\srcN}_{\srcn=1} \prod^{P}_{p=1} \prod^{N_a}_{n=1} \normal (\mathbf{Y}_{\srcn, p,n} | \frac{1}{|\res_{\srcn, n}|} \int_{\res_{\srcn, n}} \sum^{Q}_{q=1} \mathbf{W}_{p, q}(\mathbf{x}) \odot \mathbf{f}_q(\mathbf{x}) \mathop{d\mathbf{x}}, \sigma^2_\srcn \mathbf{I})^{\cweight}}_{\text{$\cmgpaggr$ Composite Likelihood}} \underbrace{\prod^{P}_{p=1} \prod^{Q}_{q=1} p(\mathbf{W}_{a,p}) \prod^{Q}_{q=1} p(\mathbf{f}_{q})}_{\text{GPRN Prior}}.  \\
\label{eq:aggr_likelihood}
\end{equation}

where $\mathbf{f}_q \sim \mathcal{N}(0, \mathbf{K}_q)$ are global functions across all tasks and $\mathbf{W}_{p,q} \sim \mathcal{N}(0, \mathbf{K}_{p,q})$ are task specific. To allow for computationally efficient inference we introduce inducing points for all latent functions  \cite{pmlr-v5-titsias09a}. For the latent basis
functions, $\mathbf{f}$, we have the inducing points $\mathbf{u} = \{ \mathbf{u}_q \}^{Q}_{q=1} ~~ \text{where} ~~ \mathbf{u}_q \in \real^{M_q}$ at locations $\mathbf{Z}^{(f)} = \{ \mathbf{Z}_q \}_{q=1}^Q ~~ \text{for} ~~ \mathbf{Z}_q \in \real^{M_q, D}$. Similarly, for the latent weight functions, $\mathbf{W}$, we
have $\mathbf{v} = \{ \mathbf{v}_{p,q} \}^{P,Q}_{p,q=1} ~~ \text{where} ~~ \mathbf{v}_{p, q} \in \real^{M_{p,q}}$ at locations $\mathbf{Z}^{(w)} = \{ \mathbf{Z}_{p,q} \}_{p,q=1}^{P,Q}$ for $\mathbf{Z}_{p,q} \in \real^{M_{p,q}, D}$. We assume that $\mathbf{f}$ and $\mathbf{W}$ are independent Gaussian processes, and
furthermore that they factor across components. We then then write the 
joint probability density including the inducing points as
\begin{equation}
\begin{aligned}
    p(\mathbf{Y},\mathbf{W},\mathbf{f},\mathbf{u},\mathbf{v}) & \propto p(\mathbf{Y}|\mathbf{W},\mathbf{f})p(\mathbf{W},\mathbf{f}|\mathbf{u},\mathbf{v})p(\mathbf{u},\mathbf{v})\\
    &= p(\mathbf{Y}|\mathbf{W},\mathbf{f}) p(\mathbf{f}|\mathbf{u}) p(\mathbf{W}|\mathbf{v})p(\mathbf{u}) p(\mathbf{v})\\
    & = p(\mathbf{Y} | \mathbf{W}, \mathbf{f}) \prod^{P}_{p=1} \prod^{Q}_{q=1} p(\mathbf{W}_{p,q}|\mathbf{v}_{p,q})p(\mathbf{v}_{p,q}) \prod^{Q}_{q=1} p(\mathbf{f}_q|\mathbf{u}_q) p(\mathbf{u}_q)
\end{aligned}
\end{equation}
where $p(\mathbf{f}_q|\mathbf{u}_q) = \normal(\mathbf{f}_q|\boldsymbol{\mu}'^{(u)}_{q}, \boldsymbol{\Sigma}'^{(u)}_{q})$ and $p(\mathbf{W}_{p,q}|\mathbf{v}_{p,q}) = \normal(\mathbf{W}_{p,q}| \boldsymbol{\mu}'^{(v)}_{p,q}, \boldsymbol{\Sigma}'^{(v)}_{p,q})$ have the
standard conditional Gaussian distributions with parameters
\begin{equation}
\begin{aligned}
	\boldsymbol{\mu}'^{(u)}_{q} &= \mathbf{K}^{(u)}_{q}(\mathbf{X}, \mathbf{Z}_{q})\mathbf{K}^{(u)}_{q}(\mathbf{Z}_q, \mathbf{Z}_{q})^{-1} \mathbf{u}_{q} &\\
	\boldsymbol{\mu}'^{(v)}_{p,q} &= \mathbf{K}^{(v)}_{p,q}(\mathbf{X}, \mathbf{Z}_{p,q})\mathbf{K}^{(v)}_{p,q}(\mathbf{Z}_{p,q}, \mathbf{Z}_{p,q})^{-1} \mathbf{v}_{p,q} &\\
	\boldsymbol{\Sigma'^{(\cdot)}_{\cdot,\cdot}} &= \mathbf{K}^{(\cdot)}_{\cdot,\cdot}(\mathbf{X}, \mathbf{X}) - \mathbf{K}^{(\cdot)}_{\cdot,\cdot}(\mathbf{X}, \mathbf{Z}_{\cdot,\cdot})\mathbf{K}^{(\cdot)}_{\cdot,\cdot}(\mathbf{Z}_{\cdot, \cdot}, \mathbf{Z}_{\cdot,\cdot})^{-1}\mathbf{K}^{(\cdot)}_{\cdot,\cdot}(\mathbf{Z}_{\cdot,\cdot},\mathbf{X}) &\\
\end{aligned}
\end{equation}
and $p(\mathbf{u}_q) = \normal(\mathbf{u}_q| 0, \mathbf{K}^{(u)}_{q})$ and $p(\mathbf{v}_{p,q}) = \normal(\mathbf{v}_{p,q}| 0, \mathbf{K}^{(v)}_{p,q})$.

\subsection{Approximate Posterior}
Variational inference turns the Bayesian posterior inference problem
into an optimisation problem where the objective function is the 
evidence lower bound (ELBO). Following \cite{Hensman:2013:GPB:3023638.3023667} 
we construct our variational lower bound, keeping $\mathbf{u}$ and  
$\mathbf{v}$ explicit, allowing for stochastic variational inference. 
We define our approximate posterior to have the factorisation
\begin{equation}
	q(\mathbf{u}, \mathbf{v}, \mathbf{f}, \mathbf{W}) = p(\mathbf{f}, \mathbf{W} | \mathbf{u}, \mathbf{v}) q(\mathbf{u}, \mathbf{v})
\end{equation}
where we have defined $q(\mathbf{u}, \mathbf{v})$ to be a free-form mixture 
of Gaussians
\begin{equation}
	q(\mathbf{u}, \mathbf{v}) =\sum^K_{k=1} \pi_k \prod^Q_{j=1} \normal(\mathbf{u}_j | \mathbf{m}^{(\mathbf{u})}_{k,j}, \mathbf{S}^{(\mathbf{u})}_{k,j}) \cdot \prod^{P, Q}_{i, j=1} \normal(\mathbf{v}_{i,j}|\mathbf{m}^{(\mathbf{v})}_{k,i, j}, \mathbf{S}^{(\mathbf{v})}_{k,i, j})	
\end{equation}
where each mean $\mathbf{m}_k \in \real^{M}$ and covariance factor 
$\mathbf{S}_k \in \real^{M \times M}$.  The marginal $q(\mathbf{f}, \mathbf{W})$ is then derived as
\begin{equation}
	\begin{aligned}
		q(\mathbf{f}, \mathbf{W}) &= \int p(\mathbf{f}, \mathbf{W} | \mathbf{u}, \mathbf{v}) q(\mathbf{u}, \mathbf{v}) \mathop{d\mathbf{u}}\mathop{d\mathbf{v}} &\\
		&= \sum^K_{k=1} \pi_k \int \prod^Q_{q=1} p(\mathbf{f}_q|\mathbf{u}_{k,q}) q(\mathbf{u}_{k,q}) \mathop{d\mathbf{u}_{k,q}} \cdot \int \prod^{P,Q}_{p,q=1} p(\mathbf{W}_{p,q}|\mathbf{v}_{k,p,q}) q(\mathbf{v}_{k,p,q}) \mathop{d\mathbf{v}_{k,p,q}}&\\
		&= \sum^K_{k=1} \pi_k \prod^Q_{q=1} \normal(\mathbf{f}_q| \boldsymbol{\mu}^{(f)}_{k,q}, \boldsymbol{\Sigma}^{(f)}_{k,q}) \cdot \prod^{P,Q}_{p,q=1} \normal(\mathbf{W}_{p,q}|\boldsymbol{\mu}^{(W)}_{k,p,q}, \boldsymbol{\Sigma}^{(W)}_{k,p,q})&\\
	\end{aligned}	
\end{equation}
where the moments of the variational distribution are given by
\begin{equation}
	\begin{aligned}
		\boldsymbol{\mu}^{(\cdot)}_{k,\cdot,\cdot} &= \mathbf{K}^{(\cdot)}_{\cdot}(\mathbf{X}, \mathbf{Z}_{\cdot})\mathbf{K}^{(\cdot)}_{\cdot}(\mathbf{Z}_\cdot, \mathbf{Z}_{\cdot})^{-1} \mathbf{m}^{(\cdot)}_{k,\cdot} &\\
		\boldsymbol{\Sigma}^{(\cdot)}_{k,\cdot,\cdot} &=  \boldsymbol{\Sigma}'^{(\cdot)}_{k,\cdot, \cdot} +  
		\mathbf{K}^{(\cdot)}_{\cdot}(\mathbf{X}, \mathbf{Z}_{\cdot})\mathbf{K}^{(\cdot)}_{\cdot}(\mathbf{Z}_\cdot, \mathbf{Z}_{\cdot})^{-1}\mathbf{S}^{(\cdot)}_{k,\cdot, \cdot}\mathbf{K}^{(\cdot)}_{\cdot}(\mathbf{Z}_\cdot, \mathbf{Z}_{\cdot})^{-1} \mathbf{K}^{(\cdot)}_{\cdot}(\mathbf{Z}_{\cdot},\mathbf{X})&\\
	\end{aligned}
\end{equation}

\subsection{Variational Lower Bound}
Following \citep{autogp_bonilla} we derive the ELBO as:

\begin{equation}
	\begin{aligned}
		\mathcal{L}_{\cmgpaggr} &= \expected_{q} \left[\log \frac{p(\mathbf{Y}, \mathbf{W}, \mathbf{f}, \mathbf{u}, \mathbf{v})}{q(\mathbf{W}, \mathbf{f}, \mathbf{u}, \mathbf{v}))}\right]\\
		& =  \expected_{q} \left[\log \frac{p(\mathbf{Y} |\mathbf{W}, \mathbf{f})p(\mathbf{f}| \mathbf{u})p(\mathbf{u}) p(\mathbf{W}|\mathbf{v})p(\mathbf{v})}{p(\mathbf{f}| \mathbf{u}) p(\mathbf{W}|\mathbf{v})q(\mathbf{u}, \mathbf{v}))}\right] &\\
		&= \underbrace{\expected_{q(\mathbf{f}, \mathbf{W})} \left[ \log p(\mathbf{Y}|\mathbf{f}, \mathbf{W})\right]}_{\text{ELL}} + \underbrace{\expected_{q(\mathbf{u},\mathbf{v})} \left[ \log \frac{p(\mathbf{u},\mathbf{v})}{q(\mathbf{u},\mathbf{v})}\right]}_{KL}
	\end{aligned}
\label{eqn:cmgp_elbo}
\end{equation}
The subsequent sections derive the closed-form expressions of both the expected log likelihood (ELL) and the KL term.
\subsection{\cmgpaggr: KL Term}

Following \cite{autogp_bonilla} the KL term is decomposed into two terms:

\begin{equation}
\begin{aligned}
	\text{KL} = \expected_q \left[ \log \frac{p(\mathbf{u}, \mathbf{v})}{q(\mathbf{u}, \mathbf{v})}\right] = \underbrace{\expected_q \left[ \log p(\mathbf{u}, \mathbf{v})\right]}_{\text{cross}} - \underbrace{\expected_q \left[ \log q(\mathbf{u}, \mathbf{v}) \right]}_{\text{ent}}
\end{aligned}
\end{equation}

where we deal with each term separately.

\subsubsection{Cross Term}

The cross term is calculated as

\begin{equation}
\begin{aligned}
	\text{cross} &= \expected_q \left[ \log p(\mathbf{u}, \mathbf{v})\right] = \sum^K_{k=1} \pi_k \expected_{q_k(\mathbf{u}, \mathbf{v})} \left[ \log p(\mathbf{u}, \mathbf{v})\right] &\\ 
	&= \sum^K_{k=1} \pi_k \left( \sum^Q_{q=1} \expected_{q(\mathbf{u}_{k,q})} \left[ \log p(\mathbf{u}_q)\right]  + \sum^P_{p=1} \sum^Q_{q=1} \expected_{q(\mathbf{v}_{k,p,q})} \left[ \log p(\mathbf{v}_{p,q})\right]\right)&\\
\end{aligned}
\end{equation}
where each
\begin{equation}
\begin{aligned}
	\expected_{q(\mathbf{u}_{k,q})}\left[ \log p(\mathbf{u}_q)\right] &= C_u - \frac{1}{2} (\mathbf{m}^{(m)}_{k,q})^T \mathbf{K}^{u}_{q}(\mathbf{Z}^{(u)}_{q}, \mathbf{Z}^{(u)}_{q})^{-1} \mathbf{m}^{(u)}_{k,q} - \frac{1}{2}Tr\left[ \mathbf{K}^{u}_{q}(\mathbf{Z}^{(u)}_{q}, \mathbf{Z}^{(u)}_{q})^{-1} \mathbf{S}^{u}_{k,q}\right] \\
	\expected_{q(\mathbf{v}_{k,p,q})}  \left[ \log p(\mathbf{v}_{p,q})\right] &= C_v - \frac{1}{2} (\mathbf{m}^{(v)}_{k,p,q})^T \mathbf{K}^{v}_{p,q}(\mathbf{Z}^{(v)}_{p,q}, \mathbf{Z}^{(v)}_{p,q})^{-1} \mathbf{m}^{(v)}_{k,p,q} - \frac{1}{2} Tr\left[ \mathbf{K}^{v}_{p,q}(\mathbf{Z}^{(v)}_{p,q}, \mathbf{Z}^{(v)}_{p,q})^{-1} \mathbf{S}^{v}_{k,p,q}\right] \\
\end{aligned}
\end{equation}
and
\begin{equation}
\begin{aligned}
	C_u &= - \frac{M_{q}}{2}\log(2\pi) - \frac{1}{2}\log |\mathbf{K}^{(u)}_{q}(\mathbf{Z^{(u)}_{q}}, \mathbf{Z^{(u)}_{q}})| &\\
	C_v &= - \frac{M_{p,q}}{2}\log(2\pi) - \frac{1}{2}\log |\mathbf{K}^{(v)}_{p,q}(\mathbf{Z^{(v)}_{p,q}}, \mathbf{Z^{(v)}_{p,q}})| &\\
\end{aligned}	
\end{equation}

\subsubsection{Entropy Term}

Following \cite{autogp_bonilla} we lower bound the entropy term of the
mixture of Gaussians as
\begin{equation}
\begin{aligned}
	\text{ent} &= \sum^K_{k=1} \pi_k \expected_{q_k(\mathbf{u}, \mathbf{v})} \left[ \log q(\mathbf{u}, \mathbf{v}) \right] &\\
	&\geq \sum^K_{k=1} \pi_k \log \left( \expected_{q_k(\mathbf{u}, \mathbf{v})} \left[  q(\mathbf{u}, \mathbf{v}) \right] \right) &\\
	&= \sum^K_{k=1} \pi_k \log \left( \sum^K_{l=1} \pi_l \expected_{q_k(\mathbf{u})} [q_l(\mathbf{u})] \cdot \expected_{q_k(\mathbf{v})} [q_l(\mathbf{v})] \right) &\\
	&= \sum^K_{k=1} \pi_k \log \left( \sum^K_{l=1} \pi_l \normal(\mathbf{m}^{(u)}_{k} | \mathbf{m}^{(u)}_{l}, \mathbf{S}^{(u)}_{k}+\mathbf{S}^{(u)}_{l}) \cdot \normal(\mathbf{m}^{(v)}_{k} | \mathbf{m}^{(v)}_{l}, \mathbf{S}^{(v)}_{k}+\mathbf{S}^{(v)}_{l})\right).
\end{aligned}	 
\end{equation}

\subsection{\cmgpaggr: Closed Form Expected Log Likelihood}

We now derive the closed form expected log likelihood (ELL) in Eq. \ref{eqn:cmgp_elbo}. The ELL is
\begin{equation}
   \text{ELL} = \sum^\srcN_{\srcn=1} \sum^P_{p=1} \sum^{N_a}_{n=1} \sum^{K}_{k=1} \pi_k \underbrace{\expected_{q_k(\mathbf{f}, \mathbf{W})} \left[ \log \normal (\mathbf{Y}_{a,p,n} | \frac{1}{|\mathcal{S}_{a,n}|} \sum_{\mathbf{x} \in \mathcal{S}_{a,n}} \sum^Q_{q=1} \mathbf{W}_{p,q}(\mathbf{x})\mathbf{f}_q(\mathbf{x}), \sigma^2_\srcn) \right]}_{\text{ELL}_a}
\label{eq:cl_likelihood_1}
\end{equation}
where each of the components can now be dealt with separately. Dealing with component $\srcn$

\begin{equation}
\begin{aligned}
	\text{ELL}_a  &=\expected_{q_k(\mathbf{f}, \mathbf{W})} \left[ \log \normal (\mathbf{Y}_{a,p,n} | \frac{1}{|\mathcal{S}_{a,n}|} \sum_{\mathbf{x} \in \mathcal{S}_{a,n}} \sum^Q_{q=1} \mathbf{W}_{p,q}(\mathbf{x})\mathbf{f}_q(\mathbf{x}), \sigma^2_\srcn) \right] &\\
	&= C_1 + C_2 \expected_{q_k(\mathbf{f}, \mathbf{W})} \left[  (\mathbf{Y}_{a,p,n}-\mu_{y})^T(\mathbf{Y}_{a,p,n}-\mu_{y}) \right] &\\
	&= C_1 + C_2 \left( \expected_{q_k(\mathbf{f}, \mathbf{W})} \left[ \mathbf{Y}_{a,p,n}^T \mathbf{Y}_{a,p,n} \right] - 2 \cdot \expected_{q_k(\mathbf{f}, \mathbf{W})} \left[ \mathbf{Y}_{a,p,n}\mu_{y} \right] + \expected_{q_k(\mathbf{f}, \mathbf{W})}  \left[\mu_{y}^2 \right]\right)&\\
	\end{aligned}	
\label{eqn:mr_gprn_ell_component}
\end{equation}
where 
\begin{equation}
	\begin{aligned}
		\mu_{y} &= \frac{1}{|\mathcal{S}_{a,n}|} \sum_{\mathbf{x} \in \mathcal{S}_{a,n}} \sum^Q_{q=1} \mathbf{W}_{p,q}(\mathbf{x})\mathbf{f}_q(\mathbf{x})&\\ 
		C_1 &=-\frac{N_a}{2}\log(2\pi \sigma^2_{a,p}) &\\
		C_2 &=-\frac{1}{2\sigma^2_{a,p}} &
	\end{aligned}
\end{equation}
and we now deal with each of these expectations separately.

\subsubsection{ELL: 1st Term}
\label{sec:exp_t1}

The first expectation does not contain $\mathbf{f}$ or $\mathbf{W}$ and so the expectations can be dropped

\begin{equation}
    \expected_{q_k(\mathbf{f}, \mathbf{W})}  \left[ \mathbf{Y}_{a,p,n}^2 \right] =  \mathbf{Y}_{a,p,n}^2
\end{equation}

\subsubsection{ELL: 2nd Term}
\label{sec:exp_t2}

For the second term the expectation is brought inside the sum, and 
then applied to each of $\mathbf{f}$ and $\mathbf{W}$ separately

\begin{equation}
\begin{aligned}
	\expected_{q_k(\mathbf{f}, \mathbf{W})}  \left[\mu_{y} \mathbf{Y}_{a,p,n} \right] &= \left( \frac{1}{|\mathcal{S}_{a,n}|} \sum_{\mathbf{x} \in \mathcal{S}_{a,n}} \sum^Q_{q=1} \expected_q \left[\mathbf{W}_{p,q}(\mathbf{x}) \right] \expected_q\left[\mathbf{f}_q(\mathbf{x})\right] \right)  \mathbf{Y}_{a,p,n} &\\  
	&= \left( \frac{1}{|\mathcal{S}_{a,n}|} \sum_{\mathbf{x} \in \mathcal{S}_{a,n}} \sum^Q_{q=1} \boldsymbol{\mu}^{(w)}_{k,p,q}(\mathbf{x}) \boldsymbol{\mu}^{(f)}_{k,q}(\mathbf{x})  \right) \mathbf{Y}_{a,p,n} &\\  
\end{aligned} 
\end{equation}
where we have used the independence property, $q(\mathbf{f}, \mathbf{W}) = 
q(\mathbf{f})q(\mathbf{W})$.
\subsubsection{ELL: 3rd Term}
\label{sec:exp_t4}
In the last expectation we have a product of sums, that is then expanded 
into a quadruple sum over $q_1$, $q_2$. There will be two cases; the first is when $q_1 = q_2$ inside the sum $\mathbf{f}$ and $\mathbf{W}$ will appear in square forms, and the second when the 
expectations can be treated as in Sec. \ref{sec:exp_t2}. 

The final term obtained after expanding the quadratic in 
\eqref{eqn:mr_gprn_ell_component} is given by
\begin{equation}
	\expected_{q_k(\mathbf{f}, \mathbf{W})}  \left[ \mu_{y}^2 \right] = \frac{1}{|\mathcal{S}_{a,n}|^2} \sum^Q_{q_1=1} \sum^Q_{q_2=1} \sum^{\mathcal{S}_{a,n}}_{\mathbf{x}_1} \sum^{\mathcal{S}_{a,n}}_{\mathbf{x}_2} \expected_{q_k(\mathbf{f}, \mathbf{W})} \left[ \mathbf{f}_{q_1}(\mathbf{x}_1) \mathbf{W}_{p,{q_1}}(\mathbf{x}_1) \mathbf{W}_{p,{q_2}}(\mathbf{x}_2) \mathbf{f}_{q_2}(\mathbf{x}_2) \right]
\end{equation}
For the case where $q_1=q_2$, then upon taking expectations we have
\begin{equation}
\begin{aligned}
	\expected_{q_k(\mathbf{f}, \mathbf{W})} & \big[ (\mathbf{f}_{q_1}(\mathbf{x}_1)\mathbf{W}_{p,q_1}(\mathbf{x}_1) \mathbf{W}_{p,q_1}(\mathbf{x}_2)\mathbf{f}_{q_1}(\mathbf{x}_2) \big] \\
	= &\expected_{q_k(\mathbf{f})}\left[ \mathbf{f}_{q_1}(\mathbf{x}_1) \left( \boldsymbol{\mu}^{(w)}_{k,p,q_1}(\mathbf{x}_1)\boldsymbol{\mu}^{(w)}_{k,p,q_1}(\mathbf{x}_2) + \boldsymbol{\Sigma}^{(w)}_{k,p,q_1} (\mathbf{x}_1,\mathbf{x}_2)\right) \mathbf{f}_{q_1}(\mathbf{x}_2)\right] &\\
	= &\boldsymbol{\Sigma}^{(w)}_{k,p,q_1}(\mathbf{x}_1,\mathbf{x}_2)\boldsymbol{\Sigma}^{(f)}_{k,q_1}(\mathbf{x}_1,\mathbf{x}_2) + \boldsymbol{\mu}^{(f)}_{k,q_1}(\mathbf{x}_1)\boldsymbol{\Sigma}^{(w)}_{k,p,q_1}(\mathbf{x}_1,\mathbf{x}_2)\boldsymbol{\mu}^{(f)}_{k,q_1} (\mathbf{x}_2)+  &\\
	&\boldsymbol{\mu}^{(w)}_{k,p,q_1}(\mathbf{x}_1)\boldsymbol{\Sigma}^{(f)}_{k,q_1}(\mathbf{x}_1,\mathbf{x}_2)\boldsymbol{\mu}^{(w)}_{k,p,q_1}(\mathbf{x}_2) + \boldsymbol{\mu}^{(f)}_{k,q_1}(\mathbf{x}_1)\boldsymbol{\mu}^{(w)}_{k,p,q_1}(\mathbf{x}_1)\boldsymbol{\mu}^{(w)}_{k,p,q_1}(\mathbf{x}_2)\boldsymbol{\mu}^{(f)}_{k,q_1}(\mathbf{x}_2)&\\
\end{aligned}
\end{equation}
where we use the notation 
$\boldsymbol{\Sigma}^{\cdot}_{\cdot, \cdot, \cdot}(\mathbf{x}_1, \mathbf{x}_2)$
to denote entry of the covariance matrix which agrees with the enumeration of
$\mathbf{x}_1$ and $\mathbf{x}_2$. And similarly $\boldsymbol{\mu}^{(w)}_{k,p,q_1}(\mathbf{x}_1) ,\boldsymbol{\mu}^{(\cdot)}_{k,\cdot}(\mathbf{x}_1) $ denotes the entry of the mean vector
agreeing with the enumeration of $\mathbf{x}_1$.
 
When $q_1 \neq q_2$ there will be no square terms and so the expectation will simply be
\begin{equation}
\expected_{q_k(\mathbf{f}, \mathbf{W})} \left[ \mathbf{f}_{q1}(\mathbf{x}_1)\mathbf{W}_{p,q1}(\mathbf{x}_1) \mathbf{W}_{p,q2}(\mathbf{x}_2)\mathbf{f}_{q2}(\mathbf{x}_2) \right] = \boldsymbol{\mu}^{(f)}_{k,q_1}(\mathbf{x}_1)\boldsymbol{\mu}^{(w)}_{k,p,q_1}(\mathbf{x}_1)\boldsymbol{\mu}^{(w)}_{k,p,q_1}(\mathbf{x}_2)\boldsymbol{\mu}^{(f)}_{k,q_1}(\mathbf{x}_2)
\end{equation}

The complete third term can now be rewritten as
\begin{equation}
\begin{aligned}
	\expected_q \left[ \mu_y^2 \right] = &\frac{1}{|\mathcal{S}_{a,n}|^2} \sum^Q_{q_1=1} \sum^Q_{q_2=1} \sum^{\mathcal{S}_{a,n}}_{\mathbf{x}_1} \sum^{\mathcal{S}_{a,n}}_{\mathbf{x}_2} \boldsymbol{\mu}^{(f)}_{k,q_1}(\mathbf{x}_1)\boldsymbol{\mu}^{(w)}_{k,p,q_1}(\mathbf{x}_1)\boldsymbol{\mu}^{(w)}_{k,p,q_1}(\mathbf{x}_2)\boldsymbol{\mu}^{(f)}_{k,q_1}(\mathbf{x}_2) + &\\ 
	&\frac{1}{|\mathcal{S}_{a,n}|^2} \sum^Q_{q=1} \sum^{\mathcal{S}_{a,n}}_{\mathbf{x}_1} \sum^{\mathcal{S}_{a,n}}_{\mathbf{x}_2} \boldsymbol{\Sigma}^{(w)}_{k,p,q}(\mathbf{x}_1,\mathbf{x}_2) \boldsymbol{\Sigma}^{(f)}_{k,q}(\mathbf{x}_1,\mathbf{x}_2) &\\ & +  \boldsymbol{\mu}^{(f)}_{k,q}(\mathbf{x_1})\boldsymbol{\Sigma}^{(w)}_{k,p,q}(\mathbf{x}_1,\mathbf{x}_2)\boldsymbol{\mu}^{(f)}_{k,q} (\mathbf{x}_2)+  
	\boldsymbol{\mu}^{(w)}_{k,p,q}(\mathbf{x}_1)\boldsymbol{\Sigma}^{(f)}_{k,q}(\mathbf{x}_1,\mathbf{x}_2)\boldsymbol{\mu}^{(w)}_{k,p,q}(\mathbf{x}_2)
\end{aligned}
\end{equation}

\subsubsection{Full Term}
Combining the derivations above we have 
\begin{equation}
\begin{aligned} 
    &\text{ELL} = 
    \sum_{\srcn=1}^{\srcN} \sum_{k=1}^{K} \pi_k
    \frac{1}{2\sigma_a^2}
    \Bigg[
    \mathbf{Y}_{a, p, n}^2 - 2 \left( \frac{1}{|\mathcal{S}_{a,n}|} \sum_{\mathbf{x} \in \mathcal{S}_{a,n}} \sum^Q_{q=1} \boldsymbol{\mu}^{(w)}_{k,p,q}(\mathbf{x}) \boldsymbol{\mu}^{(f)}_{k,q}(\mathbf{x})  \right)  \mathbf{Y}_{a,p,n} \\
    &\qquad + 
    \frac{1}{|\mathcal{S}_{a,n}|^2} \Bigg( \sum^Q_{q_1, q_2 =1}  \sum^{\mathcal{S}_{a,n}}_{\mathbf{x}_1, \mathbf{x}_2}  \boldsymbol{\mu}^{(f)}_{k,q_1}(\mathbf{x}_1)\boldsymbol{\mu}^{(w)}_{k,p,q_1}(\mathbf{x}_1)\boldsymbol{\mu}^{(w)}_{k,p,q_1}(\mathbf{x}_2)\boldsymbol{\mu}^{(f)}_{k,q_1}(\mathbf{x}_2)
    \notag  \\
    &\qquad + 
    \delta_{q_1, q_2} \cdot \bigg(
    \boldsymbol{\Sigma}^{(w)}_{k,p,q}(\mathbf{x}_1,\mathbf{x}_2)\boldsymbol{\Sigma}^{(f)}_{k,q}(\mathbf{x}_1,\mathbf{x}_2) +
    \boldsymbol{\mu}^{(f)}_{k,q}(\mathbf{x}_1)\boldsymbol{\Sigma}^{(w)}_{k,p,q}(\mathbf{x}_1,\mathbf{x}_2)\boldsymbol{\mu}^{(f)}_{k,q} (\mathbf{x}_2) 
	\boldsymbol{\mu}^{(w)}_{k,p,q}(\mathbf{x}_1)\boldsymbol{\Sigma}^{(f)}_{k,q}(\mathbf{x}_1,\mathbf{x}_2)\boldsymbol{\mu}^{(w)}_{k,p,q}(\mathbf{x}_2)
    \Bigg)\Bigg] \\
    &\qquad - \frac{1}{2}\sum_{a=1}^{\srcN} N_a P \log (2\pi \sigma_a^2).
\end{aligned}
\end{equation}
or
\begin{equation}
\begin{aligned}
&\text{ELL} = 
\sum_{\srcn=1}^{\srcN}\sum_{p=1}^P \sum_{n=1}^{N_a} \sum_{k=1}^K \pi_k
\log \mathcal{N}\left(Y_{a, p, n}  \mid 
 \frac{1}{|\mathcal{S}_{a,n}|} \sum_{\mathbf{x} \in \mathcal{S}_{a,n}} \sum^Q_{q=1} \boldsymbol{\mu}^{(w)}_{k,p,q}(\mathbf{x}) \boldsymbol{\mu}^{(f)}_{k,q}(\mathbf{x})
, \sigma_{a, p}^2
\right) \\
&\qquad - \sum_{\srcn=1}^{\srcN}\sum_{p=1}^P \sum_{n=1}^{N_a} \sum_{k=1}^K 
\frac{\pi_k}{2\sigma_{a, p}^2}
\frac{1}{|S_{a, n}|^2}
\sum_{q=1}^Q \sum_{\mathbf{x}_1, \mathbf{x}_2}
\boldsymbol{\Sigma}^{(w)}_{k,p,q}(\mathbf{x}_1,\mathbf{x}_2)\boldsymbol{\Sigma}^{(f)}_{k,q}(\mathbf{x}_1,\mathbf{x}_2)  &\\
    & + \boldsymbol{\mu}^{(f)}_{k,q}(\mathbf{x}_1)\boldsymbol{\Sigma}^{(w)}_{k,p,q}(\mathbf{x}_1,\mathbf{x}_2)\boldsymbol{\mu}^{(f)}_{k,q} (\mathbf{x}_2) 
	\boldsymbol{\mu}^{(w)}_{k,p,q}(\mathbf{x}_1)\boldsymbol{\Sigma}^{(f)}_{k,q}(\mathbf{x}_1,\mathbf{x}_2)\boldsymbol{\mu}^{(w)}_{k,p,q}(\mathbf{x}_2) &
\end{aligned}
\end{equation}

\subsection{\cmgpaggr: Closed Form Expected Log Likelihood
$\left(\mathbf{W}\rightarrow\exp(\mathbf{W})\right)$}
\label{sec:ell_pos_w}

\subsubsection{Moments of exponentiated Gaussian random variables}
\label{sec:expon_gauss}
In this section we provide additional results needed for the calculations in 
this section using standard properties of the moment generating function
of a Gaussian random variable.

\begin{equation}
   \begin{aligned}
      \expected[\exp(t \cdot \mathbf{W}_{p,q})] &= \int_{\real} \normal(\mathbf{W}_{p,q}\mid \mu_{\mathbf{W}_{p,q}}, \Sigma_{\mathbf{W}_{p,q}}) \exp(t \cdot \mathbf{W}_{p,q}) 
      \mathop{d\mathbf{W}_{p,q}} \\
        &=  \exp(t\mu_{\mathbf{W}_{p,q}} + \frac{t^2}{2} \Sigma_{\mathbf{W}_{p,q}} )  
   \end{aligned}
   \label{eq:expected_value_exp}
\end{equation}
where $\exp(\cdot)$ is defined as element-wise function,
for arbitrary $t \in \real$.


\subsubsection{ELL with positive weights}

If $\mathbf{W}$ is passed through an exponential function to enforce positive 
latent weights the expected log-likelihood is

\begin{equation}
   \alpha \sum^\srcN_{\srcn=1} \sum^P_{p=1} \sum^{N_a}_{n=1} \sum^{K}_{k=1} \pi_k \underbrace{\expected_{q_k(\mathbf{f}, \mathbf{W})}\left[ \log \normal (\mathbf{Y}_{a,p,n} | \frac{1}{|\mathcal{S}_{a,n}|} \sum_{\mathbf{x} \in \mathcal{S}_{a,n}} \sum^Q_{q=1} \exp(\mathbf{W}_{p,q}(\mathbf{x}))\mathbf{f}_q(\mathbf{x}), \sigma^2_\srcn) \right]}_{\text{ELL}_a}
\label{eq:cl_exp_likelihood_1}
\end{equation}

Each of the likelihood components can now be dealt with separately, and moreover has the same form as 
\eq \ref{eqn:mr_gprn_ell_component} with
\begin{equation}
	\mu_y = \frac{1}{|\mathcal{S}_{a,n}|} \sum_{\mathbf{x} \in \mathcal{S}_{a,n}} \sum^Q_{q=1} \exp(\mathbf{W}_{p,q}(\mathbf{x}))\mathbf{f}_q(\mathbf{x})
\end{equation}

As above, we now deal with the expectation of each term obtained after expanding the
quadratic separately, but using the results for the exponentiated moments in \ref{sec:expon_gauss}.
\subsubsection{ELL: 1st Term}

As in Sec. \ref{sec:exp_t1} the expectation is constant:

\begin{equation}
   \expected_{q_k(\mathbf{f}, \mathbf{W})} \left[ \mathbf{Y}_{a,p,n}^2\right] =  \mathbf{Y}_{a,p,n}^2
\end{equation}

\subsubsection{ELL: 2nd Term}
\label{sec:exp_t22}

As in Sec. \ref{sec:exp_t2} the 2nd term the expectation is brought inside the sum and applied to $\mathbf{f}$ and $\mathbf{W}$ separately. To evaluate the expectation of $\exp(\mathbf{W})$ we use the result from Eq. \ref{eq:expected_value_exp}: 

\begin{equation}
\begin{aligned}
	\expected_{q_k(\mathbf{f}, \mathbf{W})} \left[ \mu_y  \mathbf{Y}_{a,p,n} \right] &= \left( \frac{1}{|\mathcal{S}_{a,n}|} \sum_{\mathbf{x} \in \mathcal{S}_{a,n}} \sum^Q_{q=1} 
	\expected_{q_k(\mathbf{f}, \mathbf{W})}
	\left[\exp(\mathbf{W}_{p,q}(\mathbf{x})) \right] 
	\expected_{q_k(\mathbf{f}, \mathbf{W})}
	\left[\mathbf{f}_q(\mathbf{x})\right] \right)  \mathbf{Y}_{a,p,n} &\\  
	&= \left( \frac{1}{|\mathcal{S}_{a,n}|} \sum_{\mathbf{x} \in \mathcal{S}_{a,n}} \sum^Q_{q=1} \exp\left(\boldsymbol{\mu}^{(w)}_{k,p,q}(\mathbf{x}) + \frac{1}{2}\boldsymbol{\Sigma}^{(w)}_{k,p,q}(\mathbf{x},\mathbf{x})\right) \boldsymbol{\mu}^{(f)}_{k,q}(\mathbf{x})  \right)  \mathbf{Y}_{a,p,n}. &\\  
\end{aligned} 
\end{equation}

\subsubsection{ELL: 3rd Term}

This derivation closely follows that from \uSec \ref{sec:exp_t4}. The third term is given by
\begin{equation}
	\expected_{q_k(\mathbf{f}, \mathbf{W})} \left[ \mu_y^2 \right] = \frac{1}{|\mathcal{S}_{a,n}|^2} 
	\sum^Q_{q_1=1} \sum^Q_{q_2=1} 
	\sum^{\mathcal{S}_{a,n}}_{\mathbf{x}_1} 
	\sum^{\mathcal{S}_{a,n}}_{\mathbf{x}_2} 
	\expected_{q_k(\mathbf{f}, \mathbf{W})} \left[ \mathbf{f}_{q_1}(\mathbf{x}_1) \exp \left( \mathbf{W}_{p,q_1} (\mathbf{x}_1)\right) \exp \left( \mathbf{W}_{p,q_2}(\mathbf{x}_2)\right)\mathbf{f}_{q_2}(\mathbf{x}_2) \right]
\end{equation}
For the case where $q_1=q_2$
\begin{equation}
\begin{aligned}
	\expected_{q_k(\mathbf{f}, \mathbf{W})} & \big[ \mathbf{f}_{q_1}(\mathbf{x}_1) \exp \left( \mathbf{W}_{p,q_1}(\mathbf{x}_1) \right)  \exp \left( \mathbf{W}_{p,q_1}(\mathbf{x}_2) \right)\mathbf{f}_{q_1}(\mathbf{x}_2) \big] \\
	= &\expected_{q_k(\mathbf{f})}  \left[ \mathbf{f}_{q_1}(\mathbf{x}_1) \left( \boldsymbol{\widetilde{\mu}}^{(w)}_{k,p,q_1}(\mathbf{x}_1) \boldsymbol{\widetilde{\mu}}^{(w)}_{k,p,q_1}(\mathbf{x}_2) + \boldsymbol{\widetilde{\Sigma}}^{(w)}_{k,p,q} (\mathbf{x}_1,\mathbf{x}_2) \right) \mathbf{f}_{q_1
	}(\mathbf{x}_2)\right] &\\
	= &\boldsymbol{\widetilde{\Sigma}}^{(w)}_{k,p,q_1} (\mathbf{x}_1,\mathbf{x}_2)\boldsymbol{\Sigma}^{(f)}_{k,q_1}(\mathbf{x}_1,\mathbf{x}_2) + \boldsymbol{\mu}^{(f)}_{k,q_1}(\mathbf{x}_1)\boldsymbol{\widetilde{\Sigma}}^{(w)}_{k,p,q_1}(\mathbf{x}_1,\mathbf{x}_2)\boldsymbol{\mu}^{(f)}_{k,q_1}(\mathbf{x}_2) +  &\\
	& \boldsymbol{\widetilde{\mu}}^{(w)}_{k,p,q_1}(\mathbf{x}_1)\boldsymbol{\Sigma}^{(f)}_{k,q_1}(\mathbf{x}_1,\mathbf{x}_2)\boldsymbol{\widetilde{\mu}}^{(w)}_{k,p,q_1}(\mathbf{x}_2) + \boldsymbol{\mu}^{(f)}_{k,q_1}(\mathbf{x}_1)\boldsymbol{\widetilde{\mu}}^{(w)}_{k,p,q_1} (\mathbf{x}_1)\boldsymbol{\widetilde{\mu}}^{(w)}_{k,p,q_1}(\mathbf{x}_2)\boldsymbol{\mu}^{(f)}_{k,q_1}(\mathbf{x}_2)&\\
\end{aligned}
\end{equation}
where
\begin{align*}
    \boldsymbol{\widetilde{\mu}}^{(w)}_{k,p,q} (\mathbf{x}_1)= \exp(\boldsymbol{\mu}^{(w)}_{k,p,q}(\mathbf{x}_1)+\frac{1}{2}\boldsymbol{\Sigma}^{(w)}_{k,p,q}(\mathbf{x}_1,\mathbf{x}_1)),
\end{align*} 
and
\begin{align*}
\boldsymbol{\widetilde{\Sigma}}^{(w)}_{k,p,q}(\mathbf{x}_1,\mathbf{x}_2) = \exp(2(\boldsymbol{\mu}^{(w)}_{k,p,q} + \boldsymbol{\Sigma}^{(w)}_{k,p,q}(\mathbf{x}_1,\mathbf{x}_2))).
\end{align*}

In all other cases there will be no square terms and so the expectation will 
simply be
\begin{equation}
\begin{aligned}
    &\expected_{q_k(\mathbf{f}, \mathbf{W})} [ \mathbf{f}_{q1}(\mathbf{x}_1) \exp(\mathbf{W}_{p,q1}(\mathbf{x}_1)) \exp(\mathbf{W}_{p,q2}(\mathbf{x}_2)) \mathbf{f}_{q2}(\mathbf{x}_2) ] 
    \\ 
    &\qquad =\boldsymbol{\mu}^{(f)}_{k,q1}(\mathbf{x}_1)\boldsymbol{\widetilde{\mu}}^{(w)}_{k,p,q1}(\mathbf{x}_1)\boldsymbol{\widetilde{\mu}}^{(w)}_{k,p,q2}(\mathbf{x}_2)\boldsymbol{\mu}^{(f)}_{k,q2}(\mathbf{x}_2)
\end{aligned}
\end{equation}

\subsection{Prediction}

Although the full predictive distribution for \cmgpaggr is not available 
analytically, we are able to derive the first and second moments in closed form.

\subsubsection{Predictive Mean}
To calculate the first moment we use the approximate variational posterior in place of the true posterior. The predictive mean for task $p$ is given as

\begin{equation}
\begin{aligned}
	\expected[\mathbf{y}_p^* \mid \mathbf{x}^*, \mathbf{X}, \mathbf{Y}] &= \int \mathbf{y}_p^* \, p(\mathbf{y}_p^* \mid \mathbf{X}^*, \mathbf{X}, \mathbf{Y}) \mathop{d\mathbf{y}^*} &\\
	&\approx \sum^K_{k=1} \pi_k \int \mathbf{y}_p^* \, p(\mathbf{y}_p^* \mid \mathbf{W}_{k,p}^*, \mathbf{f}_k^*, \mathbf{x}^*, \mathbf{X}, \mathbf{Y}) q(\mathbf{f}^*_k)q(\mathbf{W}^*_k)\mathop{d\mathbf{y}_p^*} \mathop{d\mathbf{W}_{k,p}^*} \mathop{d\mathbf{f}_{k}^*}&\\
	&= \sum^K_{k=1} \pi_k \int \sum^Q_{q=1} \mathbf{W}^*_{p,q}\mathbf{f}^*_q q(\mathbf{f}^*_k)q(\mathbf{W}^*_k)\mathop{d\mathbf{W}_{k,p}^*} \mathop{d\mathbf{f}_k^*} &\\
	&= \sum^K_{k=1} \pi_k \sum^Q_{q=1} \boldsymbol{\mu}^{(w)}_{k,p,q}(\mathbf{x}^*) \boldsymbol{\mu}^{(f)}_{k,q}(\mathbf{x}^*)
\end{aligned}	
\label{eqn:a_mr_gprn_mean}
\end{equation}

\subsubsection{Predictive Variance}

The point wise second moment of task $p$ is given by
\begin{equation}
	\Var[\mathbf{y}^*_p] = \expected \left[ (\mathbf{y}^*_p )^2 \right] - \expected \left[ \mathbf{y}^*_p \right] \expected \left[ \mathbf{y}^*_p \right]
\end{equation} 
We have already calculated the closed form mean in \eq \ref{eqn:a_mr_gprn_mean} and the square form is given by
\begin{equation}
\begin{aligned}
	\expected \left[ (\mathbf{y}^*_p)^2 \right] &\approx \sum^K_{k=1} \pi_k \int (\mathbf{y}_p^*)^2   p(\mathbf{y}_p^* | \mathbf{W}_{k,p}^*, \mathbf{f}_k^*, \mathbf{x}^*, \mathbf{X}, \mathbf{Y}) q(\mathbf{f}^*_k)q(\mathbf{W}^*_k)\mathop{d\mathbf{y}_p^*} \mathop{d\mathbf{W}_{k,p}^*} \mathop{d\mathbf{f}_{k}^*}&\\
	&= \sum^K_{k=1} \pi_k \sigma^2_{\mathbf{y}_p}I + \sum^K_{k=1} \pi_k\expected_{q_k(\mathbf{f}^*, \mathbf{W}^*)} \left[ \left( \sum^Q_{q_1=1} \mathbf{W}_{k,p,q_1}^* \mathbf{f}_{k,q_1}^* \right) \left( \sum^Q_{q_2=1} \mathbf{W}_{k,p,q_2}^* \mathbf{f}_{k,q_2}^* \right) \right]
\end{aligned}
\end{equation}
In the case when $q_1 = q_2$ the expectation is
\begin{equation}
\begin{aligned}
	\sum^Q_{q_1=1} \expected_{q_k(\mathbf{f}^*, \mathbf{W}^*)} &  \left[ \mathbf{W}_{k,p,q}^* \mathbf{f}_{k,q}^* \mathbf{f}_{k,q}^*  \mathbf{W}_{k,p,q}^*  \right] \\
	&= \sum^Q_{q_1=1} \expected_{q_k(\mathbf{W}^*)} \left[ \mathbf{W}_{k,p,q_1}^* \left( \boldsymbol{\mu}^{(f)}_{k,q_1}(\mathbf{x}^*) \boldsymbol{\mu}^{(f)}_{k,q_1}(\mathbf{x}^*) + \boldsymbol{\Sigma}^{(f)}_{k,q_1}(\mathbf{x}^*,\mathbf{x}^*) \right)  \mathbf{W}_{k,p,q_1}^* \right]\\
	&= \sum^Q_{q_1=1} \big[ \boldsymbol{\mu}^{(W)}_{k,p,q_1}(\mathbf{x}^*) \boldsymbol{\mu}^{(f)}_{k,q_1}(\mathbf{x}^*) \boldsymbol{\mu}^{(f)}_{k,q_1}(\mathbf{x}^*) \boldsymbol{\mu}^{(W)}_{k,p,q_1}(\mathbf{x}^*) + \boldsymbol{\mu}^{(W)}_{k,p,q_1}(\mathbf{x}^*) \boldsymbol{\Sigma}^{(f)}_{k,q_1} (\mathbf{x}^*,\mathbf{x}^*) \boldsymbol{\mu}^{(W)}_{k,p,q_1}(\mathbf{x}^*)\\
	&+ \boldsymbol{\mu}^{(f)}_{k,q_1} (\mathbf{x}^*) \boldsymbol{\Sigma}^{(W)}_{k,p,q_1}(\mathbf{x}^*,\mathbf{x}^*) \boldsymbol{\mu}^{(f)}_{k,q_1}(\mathbf{x}^*) + \boldsymbol{\Sigma}^{(f)}_{k,q_1}(\mathbf{x}^*,\mathbf{x}^*) \boldsymbol{\Sigma}^{(W)}_{k,p,q_1}(\mathbf{x}^*,\mathbf{x}^*) \big]\\
\end{aligned}
\end{equation}
In all other cases there will be no square terms
\begin{equation}
\sum^Q_{q_1=1} \sum^Q_{q_2 \neq q_1}\expected_{q_k(\mathbf{f}^*, \mathbf{W}^*)}\left[ \mathbf{W}_{k,p,q_1}^* \mathbf{f}_{k,q_1}^* \mathbf{f}_{k,q_2}^*  \mathbf{W}_{k,p,q_2}^*  \right] = \sum^Q_{q_1=1} \sum^Q_{q_2 \neq q_1} \boldsymbol{\mu}^{(W)}_{k,p,q_1}(\mathbf{x}^*) \boldsymbol{\mu}^{(f)}_{k,q_1}(\mathbf{x}^*) \boldsymbol{\mu}^{(f)}_{k,q_2}(\mathbf{x}^*) \boldsymbol{\mu}^{(W)}_{k,p,q_2}(\mathbf{x}^*).
\end{equation}
Combining both cases together we can write

\begin{equation}
\begin{aligned}
\expected \left[ (\mathbf{y}^*_p)^2 \right] &= \sum^K_{k=1} \pi_k \sigma^2_{\mathbf{y}_p}I + \sum^K_{k=1} \pi_k \sum^Q_{q_1=1} [ \boldsymbol{\mu}^{(W)}_{k,p,q_1}(\mathbf{x}^*) \boldsymbol{\Sigma}^{(f)}_{k,q_1} (\mathbf{x}^*,\mathbf{x}^*) \boldsymbol{\mu}^{(W)}_{k,p,q_1}(\mathbf{x}^*) \\
&+ \boldsymbol{\mu}^{(f)}_{k,q_1}(\mathbf{x}^*) \boldsymbol{\Sigma}^{(W)}_{k,p,q_1}(\mathbf{x}^*,\mathbf{x}^*) \boldsymbol{\mu}^{(f)}_{k,q_1} (\mathbf{x}^*)+ \boldsymbol{\Sigma}^{(f)}_{k,q_1}(\mathbf{x}^*,\mathbf{x}^*) \boldsymbol{\Sigma}^{(W)}_{k,p,q_1}(\mathbf{x}^*,\mathbf{x}^*)] &\\
	&+ \sum^K_{k=1} \pi_k \sum^Q_{q_1=1} \sum^Q_{q_2=1} \boldsymbol{\mu}^{(W)}_{k,p,q_1} (\mathbf{x}^*) \boldsymbol{\mu}^{(f)}_{k,q_1}(\mathbf{x}^*) \boldsymbol{\mu}^{(f)}_{k,q_2}(\mathbf{x}^*) \boldsymbol{\mu}^{(W)}_{k,p,q_2}(\mathbf{x}^*). &\\
\end{aligned}
\end{equation}

\section{Synthetic Examples}
Apart from the variational experiments in Section 2 of the main paper, additional experiments using a Markov Chain Monte Carlo (MCMC) approach are conducted in this section. We show that when the dependency structure is lost through the product likelihood construction, the mean of the posterior distribution for the latent function will also deviate from the true one. We also demonstrate the posterior contraction and the effect of the different corrections.

\subsection{Data Generating Process}
We generate two synthetic observation processes from:\\

\begin{equation}
\begin{aligned}
y^{(1)}_i &= f(x_i) + \epsilon_1\\
y^{(2)}_j &= \frac{1}{2} \sum_{k=2j-1}^{2j} y^{(1)}_k  + \epsilon_2 
\end{aligned}
\label{eq:synthetic_data}
\end{equation}

Where $y^{(1)}_i, i=1,...,2N$ is the observed value of $f(x_i)$ with noise $\epsilon_1 \sim {\cal N}(0,\sigma_1^2)$ and $y^{(2)}_j , j = 1,...,N $  is the aggregate function of $y^{(1)}$ with noise $\epsilon_2 \sim {\cal N}(0, \sigma_2^2)$,$\sigma_1 =1, \sigma_2 = 0.1 $. We are using a $\sin$ function to generate data $ f(x_i) = 5\sin^2(x_i)$. The likelihood function with the observation processes $ \mathbf{Y_1} =\{y^{(1)}_i \}_{i =1}^{2N}$ , $ \mathbf{Y_2} =\{y^{(2)}_j \}_{j =1}^{N}$ is given by:
\begin{equation}
    L(\mathbf{Y}_1,\mathbf{Y}_2) = p(\mathbf{Y}_1|\mathbf{f}(\mathbf{x}),\sigma_1^2) p(\mathbf{Y}_2|\mathbf{Y}_1,\sigma_2^2)  \label{eq:true_likelihood}
\end{equation}

When the data from $\mathbf{Y}_1$ has the same support as the observation process $\mathbf{Y}_2$, the evidence from $\mathbf{Y}_2$ will not affect parameter estimation in the probability function $p(\mathbf{Y}_1|\mathbf{f}(\mathbf{x}),\sigma_1)$. However, when $\mathbf{Y}_2$ has different support from the observed $\mathbf{Y}_1$, the additional evidence should impact parameter inference. As $\mathbf{Y}_2$ does not depend on the latent function, this evidence will be hard to pass via the likelihood function in Eq. \ref{eq:true_likelihood}.  One way to correct for this is to introduce dependency between $\mathbf{Y}_1$ and $\mathbf{Y}_2$ through a non-parametric prior over the latent function $\mathbf{f(x)}$. \\

\subsection{Gaussian Processes: Product Likelihood}
Since the two observation processes follow the same underlying function $sin^2(x)$, we use a single Gaussian process to model the latent function $f(x)$.  We assume: 
\begin{equation}
  f(x) \sim {\cal GP}(0, k(x,x'))
\end{equation}

where $k(x,x')$ is the covariance function of $f(x)$. We are using the squared exponential kernel:
\begin{equation}
k(x,x') = A\exp(-\frac{(x-x')^2}{l})  
\end{equation}
where $A$ is the amplitude parameter and $l$ is the length scale for the kernel function. Thus, we can write down the joint distribution of $\mathbf{Y}_1$ and $\mathbf{Y}_2$  as:
\begin{equation}
p(\mathbf{Y}_1,\mathbf{Y}_2, \mathbf{f(x)} ) = p(\mathbf{f(x)}|\theta) p(\mathbf{Y}_1|\mathbf{f(x)},\sigma_1^2) p(\mathbf{Y}_2|\mathbf{Y}_1,\sigma_2^2)    
\end{equation}
 
Where $\theta$ is the hyper-parameters for the Gaussian process.
We can write down the distribution of $\mathbf{Y}_1$ and $\mathbf{Y}_2$ by marginalizing out the latent function:

\begin{equation}
  p(\mathbf{Y}_1,\mathbf{Y}_2 ) = \int  p(\mathbf{f(x)}|\theta) p(\mathbf{Y}_1|\mathbf{f(x)},\sigma_1^2) p(\mathbf{Y}_2|\mathbf{Y}_1,\sigma_2^2) d \mathbf{f(x)}   
\end{equation}

As $\mathbf{Y}_1$ has all the information of $\mathbf{f(x)}$, this integral is tractable and we can write $y^{(1)} \sim {\cal GP}( 0 , k(x,x')+\sigma_1)$. But when the aggregation function $\mathbf{Y}_2$ has additional information about the latent function, i.e. $\mathbf{Y}_1$ and $\mathbf{Y}_2$ only partially overlapping, bringing additional information from $\mathbf{Y}_2$ requires the prediction of the missing values of the corresponding $\mathbf{Y}_1$ process. This can be done in Markov Chain Monte Carlo(MCMC) setting by treating the unobserved value of $\mathbf{Y}_1$ as extra parameters. However, this increase a lot of computational complexity for the MCMC sampler. One option is to make an independence assumption for $\mathbf{Y}_1$ and $\mathbf{Y}_2$. Thus, the information in $\mathbf{Y}_2$ can affect the latent function $ \mathbf{f} $ directly.

\subsection{Gaussian Processes: Composite Likelihood}
Using the composite likelihoods, we assume each part of the likelihood is independent to each other. For the joint probability of $\mathbf{Y}_1$ and $\mathbf{Y}_2$ , we have:
\begin{equation}
  p(\mathbf{Y}_1,\mathbf{Y}_2 ) = \int  p(\mathbf{f(x)}|\theta) p(\mathbf{Y}_1|\mathbf{f(x)},\sigma_1^2) p(\mathbf{Y}_2|\mathbf{f(x)},\sigma_2^2) d \mathbf{f(x)}   
\end{equation}

Instead of assuming the conditional probability $p(\mathbf{Y}_2|\mathbf{Y}_1,\sigma_2^2)$, we are now assuming the data depends on the latent function $\mathbf{f(x)}$ directly. However, when $\mathbf{Y}_1$ and $\mathbf{Y}_2$ are different resolutions under the same support, this likelihood misspecifies the correlation and will make the inference of $f(x)$ contract into the observed mean. While this contraction actually equals to an extra bias to the data in the overlapping zone, the misspecified dependency structure will lead to an overfitting problem. This overfitting problem of product likelihoods has been studied in the information theory \citep{compositelikelihood_review, stoehr2015calibration} and the simplest way is to use an exponential weight to correct the inference:
\begin{equation}
 L(\mathbf{Y}_1,\mathbf{Y}_2 ) = \int  p(\mathbf{f(x)}|\theta) p(\mathbf{Y}_1|\mathbf{f(x)},\sigma_1^2)^{\alpha} p(\mathbf{Y}_2|\mathbf{f(x)},\sigma_2^2)^{\alpha} d \mathbf{f(x)}    
\end{equation}

where $ \alpha \in \real_{>0}$ is composite weight for the likelihood. The problem of learning the latent function becomes learning the parameters of the likelihood function and the composite weights. \\

\subsection{Composite Weights}
\label{sec:composite_weights}
The composite log likelihood function can be written as:
\begin{equation}
  \ell_c(\hat{\theta})= \sum_{i=1}^k f(\hat{\theta_i}|\mathbf{Y})
\end{equation}
where $ f(\hat{\theta_i}|Y) $ is the likelihood function of $i$-th parameter $\theta_i$ and we assume each part of the likelihood function is independent to each other. $ \hat{\theta_i}$ is the estimated value of $\theta_i$. With the observed distribution of $\mathbf{Y}$, $p_0(\mathbf{Y}|\theta_0)$ and $\theta_0$ as the true parameter value, we have:\\
\begin{alignat}
 1\ell_c'(\theta_0)  = \ell_c'(\hat{\theta})  & + (\theta_0 - \hat{\theta}) \ell_c''(\hat{\theta}) + o(n^{-1})    \\
  \hat{\theta} - \theta_0 & \rightarrow - \frac{\ell'(\theta_0)}{\ell''(\theta_0)}
\end{alignat}

Since we have $\ell'(\theta) = J(\theta)$ and $\ell''(\theta) = H(\theta)$, the variance of $\theta$ will follow the sandwich variance $H^{-1}(\theta)J(\theta)H^{-1}(\theta)$. Then, calculating the Taylor expansion for the likelihood, we have:
\begin{equation}
\ell_c(\theta_0) = \ell_c(\hat{\theta})  + (\theta_0 - \hat{\theta}) \ell_c'(\hat{\theta}) + \frac{1}{2} (\theta_0 - \hat{\theta}) \ell_c''(\hat{\theta}) (\theta_0 - \hat{\theta})^T +o(n^{-1})
\end{equation}

The expected variance from the composite likelihood model is :
\begin{equation}
\mathbb{E}_{\theta} [Var(\hat{\theta}|\mathbf{Y})] = -H(\hat{\theta}|\mathbf{Y})    
\end{equation}

Since $\hat{\theta}  \rightarrow \theta $, we need to set the variance of the estimated parameter to the asymptotic variance. Thus, we have:
\begin{equation}
    \mathbb{E}_{\theta} [\alpha Var(\hat{\theta}|\mathbf{Y})] = H^{-1}(\hat{\theta}|\mathbf{Y})J(\hat{\theta}|\mathbf{Y})H^{-1}(\hat{\theta}|\mathbf{Y})
\end{equation}
For a scalar variable, we can match the variance to the exact asymptotic vairance using a scalar number. But if the estimating variable $\theta$ is high dimensional, it's not easy to adjust the proper variance using a single weight. We could use a matrix ($\mathbf{C} \in \real^{k\times k} $) to adjust the covariance structure. In this case we would have:
\begin{equation}
    \mathbf{C}H(\hat{\theta}|\mathbf{Y})\mathbf{C}^T = H^{-1}(\hat{\theta}|\mathbf{Y})J(\hat{\theta}|\mathbf{Y})H^{-1}(\hat{\theta}|\mathbf{Y})  
\end{equation}
However, this increases the computational complexity substantially. One alternative way is to use a scalar weight to match the identities of the covariance matrix. \citet{2017arXiv170907616L} and \citet{Ribatet12bayesianinference} developed two different ways to adjust the identities of the covariance matrix:
\begin{equation} 
   \alpha_{\textrm{Ribatet}} = \frac{|\hat{\theta}|}{Tr[\mathbf{H}(\hat{\theta})^{-1} \mathbf{J}(\hat{\theta})]} ~~~,~~~ \alpha_{\textrm{Lyddon}} = \frac{ Tr[\mathbf{H}(\hat{\theta}) \mathbf{J}(\hat{\theta})^{-1} \mathbf{H}(\hat{\theta})]}{Tr[\mathbf{H}(\hat{\theta})]}.
\end{equation}
Where $\alpha_{\textrm{Ribatet}}$ considers all the information in the covariance matrix and $\alpha_{\textrm{Lyddon}} $  only matches the information in the diagonal elements.

\subsection{MCMC Composite Likelihood Experiments}

We now construct an MCMC experiment for the synthetic data using Eq. \ref{eq:synthetic_data}. Instead of sampling directly from the intractable joint distribution of $ L(\mathbf{Y_1},\mathbf{Y_2})$, we sample from the joint probability with the latent variable $ L(\mathbf{Y_1},\mathbf{Y_2},\mathbf{f(x)})$ via a Metropolis-Hastings within Gibbs sampler. We perform three block updates: on $\theta_0$ for the Gaussian process prior, $\mathbf{f(x)}$ for the latent function variables and $\mathbf{\sigma}^2 = \{\sigma_1^2,\sigma_2^2\}$ for the noise parameter. 

\begin{algorithm}[H]
   \caption{Block Metropolis-Hastings within Gibbs}
\begin{algorithmic}
   \State {\bfseries Input:} Observed datasets $\{(\mathbf{X}_s, \mathbf{Y}_s) \}^{S}_{s=1}$, 
   initial parameters $\theta_0$, \\
   \For{$i$-th iteration}
        \State Update parameter block $\theta_i$
        \Function{Block }{$\theta_i$}\
            \State 1. Sample proposed value of the Gaussian process prior $\theta'_i \sim N(\theta_{i-1}, \Delta)$ \\
            \State 2. Calculate the conditional probability distribution $p(\theta'_i|Y_1,Y_2,\mathbb{\sigma}^2_{i-1},f(x)_{i-1})$
            \State 3. Calculate the acceptance rejection ratio:\\
            \State $\pi = \frac{p(\theta'_i|Y_1,Y_2,\mathbb{\sigma}^2_{i-1},\mathbf{f(x)}_{i-1})}{p(\theta_{i-1}|Y_1,Y_2,\mathbb{\sigma}^2_{i-1},\mathbf{f(x)}_{i-1})}$\\
            \State 4. Update the $i$-th value of $\theta_i$ via $\pi$   \\
        \EndFunction\\
        \State Update the parameter block $\mathbf{f(x)}_i$ via $ \pi = \frac{p(\mathbf{f(x)}'_i|Y_1,Y_2,\mathbb{\sigma}^2_{i-1},\theta_i)}{p(\mathbf{f(x)}_{i-1}|Y_1,Y_2,\mathbb{\sigma}^2_{i-1},\theta_{i})}$\\
        \State Update the parameter block $\mathbb{\sigma}_i$ via $\pi = \frac{p(\mathbb{\sigma'}^2_i|Y_1,Y_2,\theta_i,\mathbf{f(x)}_i)}{p(\mathbb{\sigma}^2_{i-1}|Y_1,Y_2,\theta_i,\mathbf{f(x)}_i)}$ \\
   \EndFor
   
$\textbf{return} {\mathbf{\theta},\mathbf{f(x)},\mathbf{\sigma}^2}$

\end{algorithmic}
\end{algorithm}

\begin{figure}[t]
\centering
\begin{subfigure}[t]{0.5\linewidth}
  \centering
  \includegraphics[width=0.96\textwidth]{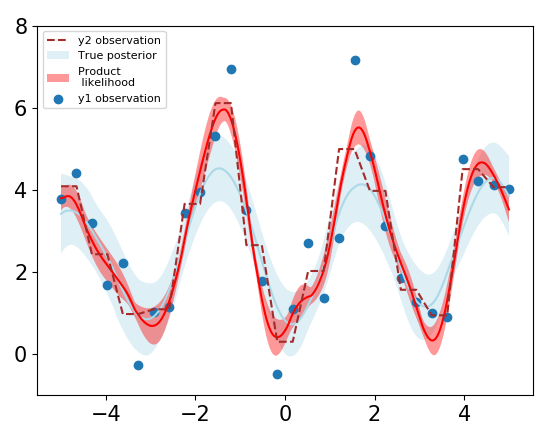}
\end{subfigure}%
~
\begin{subfigure}[t]{0.5\linewidth}
  \centering
  \includegraphics[width=\textwidth]{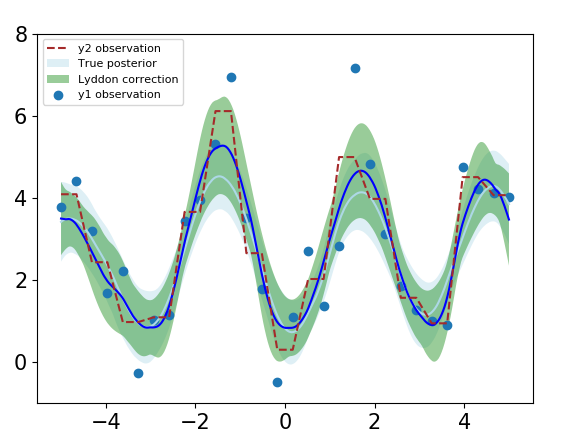}
\end{subfigure}
\begin{subfigure}[t]{0.5\linewidth}
   \centering
   \includegraphics[width=1.01\textwidth]{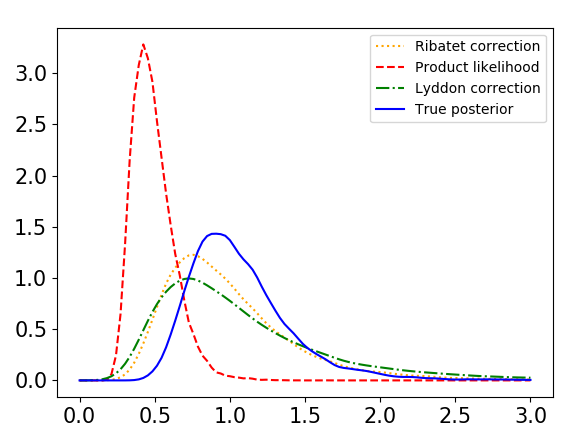}
\end{subfigure}%
~
\begin{subfigure}[t]{0.5\linewidth}
   \centering
   \includegraphics[width=\textwidth]{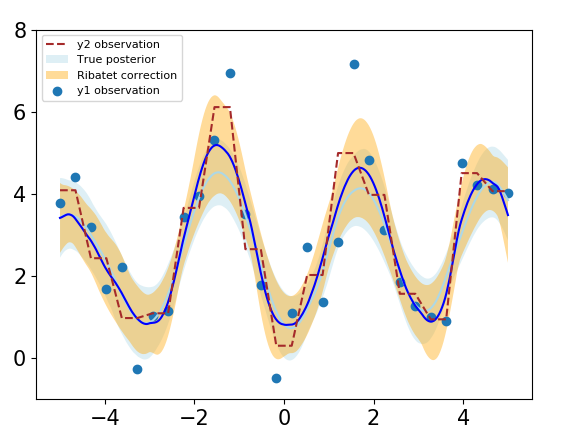}
\end{subfigure}
\caption{\textbf{Top Left:} Comparing the posterior of the latent function $f(x)$ under the product likelihood assumption and a correctly specified likelihood. The product likelihood assumption causes extreme posterior contraction which effects both the mean and variance. \textbf{Bottom Left:} Comparison of the true posterior of $\sigma_1$ noise to the posterior under the product likelihood and under the composite likelihood with different weights. The composite likelihood is able to recover the true posterior. \textbf{Top Right:} Comparing the posterior of the latent function $f(x)$ under the composite likelihood assumption with Lyddon correction and a correctly specified likelihood.\textbf{Bottom Right:} Comparing the posterior of the latent function $f(x)$ under the composite likelihood assumption with Ribatet correction and a correctly specified likelihood. The two correction have the similar results for our experiments. Although the mean of the function is not exactly match the true function, the variance of the latent function is corrected close to the true function. The misspecified part is due to the imprefect match of the asymptotic variance discussed in section \ref{sec:composite_weights}}
\end{figure}

\newpage
\subsection{Variational Composite Likelihood Experiments}

In this section we provide further details to reproduce the variational composite likelihood experiments in Sec. 2.2 and Fig. 2 of the main paper.

\textbf{Data Generation} We consider the case of having two dependent observation processes. We generate one process $\mathbf{Y}_1 = 5 \cdot \sin(\mathbf{X})^2 + 0.1\cdot \epsilon$ with $\epsilon \sim \normal(0, 1)$ with 100 samples over the range $[-2, 15]$. For $\mathbf{Y}_2$ we aggregate $\mathbf{Y}_1$ into bins of size 3, $\res_2 = 3$, so that $\mathbf{Y}_2 \in \real^{33}$ and $\mathbf{X}_2 \in \real^{ 33\times 3}$. In Fig. 2 we only plot the range $[3, 10]$. 

\textbf{Parameter Initialization} For both \cmgpgp and \vbagg we use an SE kernels with lengthscale of 0.1 and variance 1.0. We initialize the likelihood noise to 0.1.

\textbf{Additional Training Details} For \vbagg we run for 10000 epochs. For \cmgpgp for we run the MLE estimate for 10000 epochs, obtain $\alpha_{\text{Ribatet}}$ and then optimize the ELBO for 10000 epochs.

\section{Multi-resolution Air Pollution Experiments}
\subsection{Inter-task Multi-resolution: PM$_{10}$-PM$_{25}$}
In this section we provide additional details for reproducing the inter-task multi-resolution experiments as described in the main paper.

\textbf{Variational Parameter Initialization}: For \cmgpdgp we initialize all likelihood noises to 0.01 and we use a Matern32 kernel for all latent functions with a lengthscale of 0.01. For both \cmgpaggr and \centerpoint we initialize the likelihood noise to be 0.1 and use a squared exponential kernel for all latent functions. We use $Q=1$ and set the lengthscale of $\mathbf{f}$ to be 0.1 and the lengthscales of $\mathbf{W}$ to be 3.0. 

\textbf{Training Details}: We train \cmgpdgp for a total of 900 iterations. We train both \cmgpaggr and \centerpoint for 2000 iterations each.

\subsection{Intra-task Multi-resolution: Space-time NO$_{2}$}
In this section we provide additional details for reproducing the \emph{intra}-task multi-resolution experiments described in Sec. 4 of the main paper.

\textbf{Data pre-processing}: We extract spatial features based on the London road network (OS Highways) \footnote{\url{https://www.ordnancesurvey.co.uk/business-and-government/help-and-support/products/os-mastermap-highways-network.html}} and land use (UKMap) \footnote{\url{https://www.geoinformationgroup.co.uk/ukmap}}. OS Highways is a dataset of every road in London with information of the length, road classification (A Road, B Road, etc). UKMap is a dataset of polygons where each polygon represents a physical entity, e.g a building, a river, a park, etc.  UKMap provides additional information such as the height of the buildings and the area of the parks and rivers.  For each input location we construct a buffer of approximately 100m (a radius 0.001 degrees in SRID:4326). Within the buffer zone we calculate the average length of the A-roads, the average ratio between the width of the roads and height of buildings on the corresponding roads, and the total area of vegetation and water. We convert all time stamps into unix epochs and we standardize all features before training. To approximate the integral in the likelihood (Eq. 4 in main text) we discretize the area of each satellite based observation input into a 10 by 10 uniform grid of lat-lon points.

\textbf{\cmgpdgp Architecture}: For \cmgpdgp we use the architecture described on the right subfigure of Fig. 3 in the main paper where $\mathbf{X}_2, \mathbf{Y}_2$ corresponds to the LAQN dataset and $\mathbf{X}_1, \mathbf{Y}_1$ to the satellite dataset. We give the initialization of the specific latent functions below.

\textbf{Variational Parameter Initialization}: For \cmgpaggr and \vbagg we use 400 inducing points for all latent functions. Both the inducing function values and the variances are randomly initialized between 0 and 1. For \cmgpdgp the latent functions $\mathbf{f}_{2, 2}$ and $\mathbf{f}_{1, 1}$ we place 300 inducing points and for $\mathbf{f}_2$ we use 100. For all models we initialize the inducing points locations with K-means.

\textbf{Model Parameter Initialization}: In all models and latent function withing, \cmgpaggr, \vbagg and \cmgpdgp we use SE kernels  initialized with lengthscales of 0.1 and SE variance to 1.0. We initialize the likelihood noise to be 5.0. 

\textbf{Additional Training Details}: We train \cmgpdgp for a total of 1500 iterations. We train both \cmgpaggr, \vbagg and \centerpoint for 2000 iterations each. 

\section{\cmgpdgp}

In this section we provide the complete derivation of the variational lower bound for \cmgpdgp. 

\subsection{Specification of the prior}

The full prior for \cmgpdgp is given by
\begin{equation}
	p(\mathbf{F}, \mathbf{M}) = \left( \prod^P_{p=2} p(\mathbf{f}_p\mid\mathbf{m}_p)p(\mathbf{m}_p\mid \mathbf{f}_{1,p}, \{\mathbf{f}^{(2)}_{a,p} \}^{\srcN}_{\srcn=2}) \right) \left(\prod^P_{p=1} p(\mathbf{f}_{1,p}) \prod^\srcN_{a=2}  
	p(\mathbf{f}^{(2)}_{a,p} \mid \mathbf{f}_{a,p})p(\mathbf{f}_{a,p}) \right)
\end{equation}
We say that each of the GPs $\mathbf{f}_{a,p}$ are base GPs and to simplify notation in subsequent 
sections we use the function $\uParent(\cdot)$ to denote the set of parent functions for each
node. For example, $\uParent(\mathbf{m}_2) =  \{ \mathbf{f}_{1,p} \} \cup \{\mathbf{f}^{(2)}_{a,p}\}$. 
\subsection{Likelihood}

The likelihood for \cmgpdgp (as illustrated in in Fig. \ref{fig:mr_dgp_plate}) is

\begin{equation}
	p(\mathbf{Y} \mid \mathbf{F}) = \prod^P_{p=2} p(\mathbf{Y}_{1,1} \mid \mathbf{f}_p)\prod^P_{p=1} \prod^\srcN_{\srcn=1} p(\mathbf{Y}_{1,p} \mid \mathbf{f}^{(2)}_{a, 1})p(\mathbf{Y}_{a,p} | \mathbf{f}_{a,p})
\end{equation}

where each likelihood component is a multi-resolution likelihood of the form

\begin{equation}
	 \normal \left(\mathbf{Y}_{\srcn, n} \; \bigg| \; \frac{1}{|\mathcal{S}_a|} \int_{S_a} \mathbf{f}^{(\nump)}_{\srcn}(\mathbf{X}) \mathop{d \mathbf{X}} , \sigma^2_{a,k} \mathbf{I} \right),
\end{equation}

again we discretise the integral with a uniform grid over $S_a$. 

\subsection{Augmented Prior}

To allow for efficient inference we sparsify each GP by introducing inducing points:

\begin{equation}
\begin{aligned}
	p(\mathbf{F}, \mathbf{M}, \mathbf{U}) = &p(\mathbf{F}, \mathbf{M} | \mathbf{U}) p(\mathbf{U}) &\\
	= &\left( \prod^P_{p=2} p(\mathbf{f}_p|\mathbf{m}_p, \mathbf{u}_p) p(\mathbf{m}_p | \uParent(\mathbf{m}_p)) \right) \cdot &\\
	&\left(\prod^P_{p=1} p(\mathbf{f}_{1,p} | \mathbf{u}_{1,p}) \prod^\srcN_{a=2}  
	p(\mathbf{f}^{(2)}_{a,p} | \mathbf{f}_{a,p}, \mathbf{u}^{(2)}_{a,p})  p(\mathbf{f}_{a,p}| \mathbf{u}_{a,p})  \right) \cdot &\\
	& \left(\prod^{P}_{p=2} p(\mathbf{u}_p) \cdot \prod^{P}_{p=1} p(\mathbf{u}_{1,1}) \prod^\srcN_{\srcn=2} p(\mathbf{u}^{(2)}_{a,p}) p(\mathbf{u}_{a,p}) \right) &\\
\end{aligned}
\end{equation}

where each $p(\mathbf{u}^{(\cdot)}_{\cdot,\cdot}) = \normal(\mathbf{u}_{\cdot, \cdot}^{(\cdot)} \mid 0, \mathbf{K}^{(\cdot)}_{\cdot, \cdot}(\mathbf{Z}^{(\cdot)}_{\cdot, \cdot}, \mathbf{Z}^{(\cdot)}_{\cdot, \cdot}))$ for $\mathbf{u}^{(\cdot)}_{\cdot,\cdot} \in \real^{M}$. The locations of the inducing points for the base GPs are $\mathbf{Z}_{a,p} \in \real^{M \times D}$ and for the deep GPs $\mathbf{Z}_{p}, \mathbf{Z}^{2}_{a,p} \in \real^{M \times D}$. For brevity we have omited the conditional on the inducing locations $\mathbf{Z}$ in our notation.

\subsection{Variational approximate Posterior}

Following \cite{NIPS2017_7045} we construct an approximate augmented posterior that maintains the dependency structure between layers:
\begin{equation}
\begin{aligned}
	q(\mathbf{M}, \mathbf{F}, \mathbf{U}) &= p(\mathbf{M}, \mathbf{F} | \mathbf{U}) q(\mathbf{U}) &\\
	&= p(\mathbf{M}, \mathbf{F} | \mathbf{U}) \prod^{P}_{p=2} q(\mathbf{u}_p) \cdot \prod^{P}_{p=1} q(\mathbf{u}_{1,1}) \prod^\srcN_{\srcn=1} q(\mathbf{u}^{(2)}_{a,p}) q(\mathbf{u}_{a,p}) &\\ 
\end{aligned}
\end{equation}

where each $q(\mathbf{u}^{(\cdot)}_{\cdot,\cdot})$ are standard free-form Gaussians $\normal(\mathbf{u}_{\cdot, \cdot}^{(\cdot)} \mid \mathbf{m}^{(\cdot)}_{\cdot,\cdot}, \mathbf{S}^{(\cdot)}_{\cdot,\cdot})$. The conditional is of the form

\begin{equation}
	p(\mathbf{f}_{a,p} | \uParent(\mathbf{f}_{a,p}), \mathbf{u}_{a,p}) = \normal(\mathbf{f}_{a, p} \mid \mu'_{a,p}, \Sigma'_{a,p}) 
\end{equation}

with mean and variance given by the standard conditional equations
\begin{equation}
\begin{aligned}
\mu'_{a,p} &=\alpha_{a,p} \mathbf{K}_{a,p}(\mathbf{Z}_{a,p}, \mathbf{Z}_{a,p}) ^{-1} \mathbf{u}_{a,p}, &\\
	\Sigma'_{a,p} &= \mathbf{K}_{a,p}(\uParent(\mathbf{f}_{a,p}), \uParent(\mathbf{f}_{a,p})) -  \alpha_{a,p}\mathbf{K}_{a,p}(\mathbf{Z}_{a,p}, \mathbf{Z}_{a,p}) ^{-1} \alpha_{a,p}^T. &\\
\end{aligned}
\end{equation}

where $\alpha_{a,p} = \mathbf{K}_{a,p}(\uParent(\mathbf{f}_{a,p}), \mathbf{Z}_{a,p})$. 

\subsection{Marginalisation over inducing points}

Firstly we can marginalise the inducing variables analytically 

\begin{equation}
\begin{aligned}
	q(\{\mathbf{m}_p\}^P_p,\{ \mathbf{f}_{a,p}\}^{\srcN,P}_{a,p=1}) &= \int q(\{\mathbf{m}_p\}^P_p,\{ \mathbf{f}_{a,p}, \mathbf{u}_{a,p} \}^{\srcN,P}_{a,p=1}) \mathop{d\mathbf{u}_{1,1}} \cdots \mathop{d\mathbf{u}_{\srcN,P}} &\\
	&= \prod^P_{p=1} p(\mathbf{m}_p | \uParent(\mathbf{m}_p)) \prod^\srcN_{a=1} \int p(\mathbf{f}_{a,p} | \uParent(\mathbf{f}_{a,p}), \mathbf{u}_{a,p}) q(\mathbf{u}_{a,p})\mathop{d\mathbf{u}_{a,p}} &\\
	&= \prod^P_{p=1} p(\mathbf{m}_p | \uParent(\mathbf{m}_p)) \prod^\srcN_{a=1} q(\mathbf{f}_{a,p} | \uParent(\mathbf{f}_{a,p})) &\\
\end{aligned}
\end{equation}

The integral can evaluated in closed form resulting in

\begin{equation}
	q(\mathbf{f}_{a,p} | \uParent(\mathbf{f}_{a,p}))  = \normal(
	\mathbf{f}_{a, p} \mid \mu_{a,p}, \Sigma_{a,p})
\end{equation}

where the mean and variance are given by
\begin{equation}
\begin{aligned}
\mu_{a,p} &=\alpha_{a,p} \mathbf{K}_{a,p}(\mathbf{Z}_{a,p}, \mathbf{Z}_{a,p}) ^{-1} \mathbf{m}_{a,p}, &\\
	\Sigma_{a,p} &= \Sigma'_{a,p} - \alpha_{a,p}\mathbf{K}_{a,p}(\mathbf{Z}_{a,p}, \mathbf{Z}_{a,p}) ^{-1} \mathbf{S}_{a,p} \mathbf{K}_{a,p}(\mathbf{Z}_{a,p}, \mathbf{Z}_{a,p}) ^{-1} \alpha_{a,p}^T\ &\\
\end{aligned}
\end{equation}

\subsection{Mixture of Experts}

We define $p(\mathbf{m}_p | \uParent(\mathbf{m} _p))$ as a mixture of experts, that is a weighted combination of the local experts
\begin{equation}
	p(\mathbf{m}_p | \uParent(\mathbf{m}_p)) = \normal\left(\mathbf{m}_{p} \;\bigg|\;
	\sum^\srcN_\srcn \mathbf{w}_{a,p}\mu_{a,p}, \sum^\srcN_\srcn \mathbf{w}_{a,p} \Sigma_{a,p} \mathbf{w}_{a,p}\right).
\end{equation}
The weights $\mathbf{w}_{a,p}$ are application specific and we provide specific examples in \uSec \ref{sec:moew}.

\subsection{Marginalisation over layers}

We follow the doubly stochastic framework of \cite{NIPS2017_7045} and marginalise through the layers using Monte Carlo estimates. The first layer of GPs can be sampled from directly, these samples are then propagated through all the subsequent layers.

\begin{equation}
	q(\mathbf{f}_{a,p}) = \int 	q(\mathbf{f}_{a,p} | \uParent(\mathbf{f}_{a,p}))  \prod^{\numL-1}_{l=1} q(\uParent^{(l)}(\mathbf{f}_{a,p}) | \uParent^{(l+1)}(\mathbf{f}_{a,p})) \mathop{d \uParent^{(1)}(\mathbf{f}_{a,p})} \cdots \mathop{d \uParent^{(\numL)}(\mathbf{f}_{a,p})}
\end{equation}

\subsection{Variational Lower Bound}

In this section we provide the derivation of the variational lower bound for \cmgpdgp. The evidence lower bound (ELBO), which lower bounds the log marginal likelihood $\log p(\mathbf{Y} | \mathbf{X})$, is 

\begin{equation}
    \begin{aligned}
   \mathcal{L} &= \expected_{q(\mathbf{M}, \mathbf{F}, \mathbf{U})} \left[ \log \frac{p(\mathbf{Y, \mathbf{M}, \mathbf{F}, \mathbf{U}})}{q(\mathbf{M}, \mathbf{F}, \mathbf{U}) } \right]  =  \expected_{q(\mathbf{M}, \mathbf{F}, \mathbf{U})} \left[ \log \frac{p(\mathbf{Y | \mathbf{F}) p(\mathbf{M}, \mathbf{F} | \mathbf{U}}) p(\mathbf{U})}{p(\mathbf{M},\mathbf{F}|\mathbf{U})q(\mathbf{U}) } \right] &\\
&\\
\end{aligned}
\end{equation}
Cancelling the relevant terms inside the logarithm we get
\begin{alignat*}{2}
   \mathcal{L}_{\cmgpdgp} &= \underbrace{\expected_{q(\mathbf{M}, \mathbf{F}, \mathbf{U})} \left[ \log p(\mathbf{Y} | \mathbf{F}) \right]}_{\text{ELL}} + \underbrace{\expected_{q(\mathbf{U})} \left[ \log \frac{P(\mathbf{U})}{q(\mathbf{U})} \right]}_{\text{KL}}.
\end{alignat*}

Note that we have slightly abused notation to keep the derivation clear and now provide the full expanded lower bound:

\begin{equation}
	\text{ELL} = \sum^P_{p=2} \expected_{q(\mathbf{f}_p)} \left[ \log p(\mathbf{Y}_{1,1} | \mathbf{f}_p) \right]+ \sum^P_{p=1} \sum^\srcN_\srcn \left[ \expected_{q(\mathbf{f^{(2)}_{a, 1}})} \left[p(\mathbf{Y}_{1,p} | \mathbf{f^{(2)}_{a, 1}})\right] + \expected_{q(\mathbf{f}_{a,p})} \left[p(\mathbf{Y}_{a,p} | \mathbf{f}_{a,p})\right] \right]
\end{equation}

and 
\begin{equation}
	\text{KL} = \expected_{q(\mathbf{U})} \left[ \log \frac{\left(\prod^{P}_{p=2} p(\mathbf{u}_p) \cdot \prod^{P}_{p=1} p(\mathbf{u}_{1,1}) \prod^\srcN_{\srcn=2} p(\mathbf{u}^{(2)}_{a,p}) p(\mathbf{u}_{a,p}) \right)}{\left(\prod^{P}_{p=2} q(\mathbf{u}_p) \cdot \prod^{P}_{p=1} q(\mathbf{u}_{1,1}) \prod^\srcN_{\srcn=2} q(\mathbf{u}^{(2)}_{a,p}) q(\mathbf{u}_{a,p}) \right)} \right].
\end{equation}

The KL term can be computed in closed form because it is just the sum of KL terms between two Gaussians. The ELL term is approximated using the Monte Carlo estimates from marginalising through the layers.

\subsection{Mixture of Experts Weights}
\label{sec:moew}

In this section we provide a specific and intuitive example of weights used when combining the mixture of experts. We derive weights that naturally weigh the experts by the level of support provided. We assume that this is defined by the resolution of the base layers, where higher resolutions are `trusted' more. We first derive the weights for two generic GPs  $\mathbf{f}_1$ and $\mathbf{f}_2$ and will generalise and apply the weights to the $\cmgpaggr$ after.

To find the support of an expert we utilise the differential entropy of the GP (see \cite{Krause:2008:NSP:1390681.1390689}):
\begin{equation}
	H(p(\mathbf{f}| \mathbf{D})) = \frac{1}{2} \log (\sigma^2_{\mathbf{f}|\mathbf{D}}) + \frac{1}{2}(\log(2\pi)+1).
\label{eqn:gp_info_gain}
\end{equation}
For our mixing weights we are not interested in the amount of information each  GP has, just whether there is any. So we drop the 2nd term on the rhs of \eq \ref{eqn:gp_info_gain}, which we denote by $I(p(\mathbf{f}| \mathbf{D}))$, and normalise the information to be between zero and one. One such function is simply:
\begin{equation}
\hat{I}(\mathbf{f}) =  \frac{I(p(\mathbf{f}|D)) - \min( I(p(\mathbf{f}|D)))}{\max( I(p(\mathbf{f}|D)))-\min( I(p(\mathbf{f}|D)))}
\end{equation}
This is not the only normalisation that can be done, one could also use a sigmoid function such as $\tanh(\cdot)$. Now we want a function that weighs $\mathbf{f}_2$ down when $\mathbf{f}_1$ has information, inherently capturing that we trust $\mathbf{f}_1$ over $\mathbf{f}_2$. One such function is the normalised information of $\mathbf{f}_2$ minus the joint information of $\mathbf{f}_1$ and $\mathbf{f}_2$:
\begin{equation}
	\hat{I}(\mathbf{f}_1) + \beta_2\hat{I}(\mathbf{f}_2) = \hat{I}(\mathbf{f}_1) + \hat{I}(\mathbf{f}_2) - \hat{I}(\mathbf{f}_1, \mathbf{f}_2).
\end{equation}

Because we have normalised the information to be within zero and one we approximate the the joint information with a hadamard product (approximating an XOR function where the value is one if they both have information else it is zero).

\begin{equation}
\begin{aligned}
	\beta_2\hat{I}(\mathbf{f}_2) &= \hat{I}(\mathbf{f}_2) - \hat{I}(\mathbf{f}_1) \hat{I}(\mathbf{f}_2) &\\
	&= ( 1- \hat{I}(\mathbf{f}_1)).
\end{aligned}
\end{equation}

The function $\beta_2$ now has maximum and minimum values in the range $[0,1]$ and so we can directly use $\beta_2$ as our mixing weights. To makes all the weights sum to one we define
\begin{equation}
	\beta_1 = 1 - \beta_2 = \hat{I}(\mathbf{f}_1).
\end{equation}

\subsection{Combining arbitrary number of experts}

We combine them in a hierarchical manner, in a similar way to \cite{Deisenroth:2015:DGP:3045118.304527}, and construct a computational graph. Given $P$ experts, the mixing weights are defined as
\begin{equation}
	\mathbf{m} =  \hat{I}(\mathbf{f}_1) \mathbf{f}_1 + (1- \hat{I}(\mathbf{f}_1))( \hat{I}(\mathbf{f}_2) \mathbf{f_2} + (1-\hat{I}(\mathbf{f}_2))( \cdots +(1-\hat{I}(\mathbf{f}_{P-1})) \mathbf{f}_P)
\end{equation}

\section{Relation to VBAgg}

In this section we show that \cmgpaggr can be seen as a generalisation of \vbagg \cite{law2018variational} from a single GP to a GPRN. In VBAgg each observation $y^{a}$ is the aggregate output of some bag of items $\mathbf{x}^a = \{\mathbf{x}_i^a\}_{i=1}^{N_a}$. The likelihood of each bag is $y^a|\mathbf{x}^a \sim \normal(y|\eta^a, \tau^a) ~~ \text{where} ~~ \eta^a = \sum^{N_a}_{i=1} w^a_i \mu(\mathbf{x}^a_i)$ and $\mu$ is the mean of the latent process $f$. In \cmgpaggr we are modelling the underlying process with the sum of products of GPs. Rewriting \cmgpaggr using the notation of \cite{law2018variational}: $\mu = \mathbf{W}\mathbf{f}$ and each dataset $\{\mathbf{X}_\srcn, \mathbf{Y}_\srcn\}^\srcN_{\srcn=1}$ directly corresponds to the observations and bag of items defined in \vbagg. Let $N_a=\res_a$, and $\tau^a = \sigma^2_\srcn$ and the composite weight $\alpha=1$. The composite weight of value 1 is implicitly included in the model of \vbagg through the independence assumption. We assume an simple aggregation of the bag of items, although we note that is not necessary, so setting $w^a_i = \frac{1}{|\res_a|}$ we obtain $y^a \sim  \normal(\sum^{N_a}_{i=1} w^a_i \mu(\mathbf{x}^a_i), \tau^a)$ which is \cmgpaggr in the notation of \cite{law2018variational}. \vbagg is then recovered when we use only one latent function (by setting $\mathbf{W}$ to a constant value), by only considering the single task setting and by setting the composite weight to one.

The VBAgg model is defined by
\begin{equation}
    y^a \sim p(y^a|\eta^a), \quad \eta^a = \sum_{i=1}^{N_a} w_i^a \eta_i^a= \sum_{i=1}^{N_a} w_i^a \Psi(f(x_i^a ))
\end{equation}
where $y^a$ is the independent aggregate observations, $w_i^a$ are fixed weights and $f$ follows a Gaussian process. For the unobserved latent variables $y_i^a$, we assume:
\begin{equation}
    y^a_i \sim p(y^a_i|\eta^a_i), \quad \eta^a_i = \Psi(f(x_i^a ))
\end{equation}
When the probabilities of $y^a \sim p(y^a|\eta^a)$ and $y^a_i \sim p(y^a_i|\eta^a_i)$ are Gaussian, the model is equivalent to \vbagg. We firstly consider the special case, when $\Psi(f(x^a_i)) =f(x^a_i)$. As $f(x^a_i) \sim {\cal GP} (0, K(x^a_i, \cdot ))$, we could say that $y^a_i \sim N(f(x^a_i),\sigma^a_i)$ still follows a Gaussian process and $w_i y^a_i \sim {\cal GP} (0 , w_i^2 K(x^a_i, \cdot))$ with fixed weight $w_i$. Thus, due to the additive property of Gaussian processes, the aggregation function $y^a \sim {\cal GP} (0, \sum_i \sum_{.} w_iw_.K(x_i^a,\cdot))$. The covariance of different aggregated observations is:
\begin{equation}
\begin{aligned}
Cov(y^a,y^b) & = Cov \left(\sum_{i = 1}^{N_a} w^a_i y^a_i  ,\sum_{j = 1}^{N_b} w^b_j y^b_j \right) \\
& = \expected \left[ \sum_{i = 1}^{N_a} \sum_{j = 1}^{N_b}  w^a_i y^a_i  w^b_j y^b_j\right] - \expected \left[\sum_{i = 1}^{N_a} w^a_i y^a_i  \right]\expected \left[\sum_{j = 1}^{N_b} w^b_j y^b_j  \right] \\
& = \expected \left[ \sum_{i = 1}^{N_a} \sum_{j = 1}^{N_b}  w^a_i y^a_i  w^b_j y^b_j\right] \\
& = \sum_{i = 1}^{N_a} \sum_{j = 1}^{N_b} w^a_i w^b_j K(x^a_i,x^b_j)\\
& = \widetilde{K}(\mathbf{x}^a,\mathbf{x}^b)
\end{aligned}
\label{eqn:agg_correlation}
\end{equation}
Also, the covariance between $y_i^* = f(x_i^*)$ and $y^a$ is given:
\begin{equation}
\begin{aligned}
Cov(y_i^*,y^a) & = Cov \left(y_i^*  ,\sum_{j = 1}^{N_a} w^a_j y^a_j \right) \\
& = \sum_{j = 1}^{N_a} w^a_j K(x^*_i,x^a_j)\\
& = \widetilde{K^*}(x^*_i,\mathbf{x}^a)
\end{aligned}
\end{equation}
Thus, the problem of learning $f(\cdot)$ function becomes a standard Gaussian process regression problem and we have:
\begin{equation}
    \begin{aligned}
    \begin{bmatrix}
    \mathbf{y}\\
    y_i^*
    \end{bmatrix} \sim N\left( 0,     \begin{bmatrix}
    \widetilde{K}(\mathbf{x},\mathbf{x}) + \sigma^2_y \mathbf{I}& \widetilde{K^*}(x^*_i,\mathbf{x})\\
    \widetilde{K^*}(\mathbf{x},x^*_i) & K(x^*_i,x^*_i)
    \end{bmatrix} \right)
    \end{aligned}
\end{equation}
where $\mathbf{y} =\{ y^1,...,y^a,...,y^n\}$ and the predictive posterior is given by: 
\begin{equation}
\begin{aligned}
    y_i^* | \mathbf{y},\mathbf{x},x_i^* & \sim N(\mu^*,\sigma^*)\\
    \mu* &= \widetilde{K^*}(x^*_i,\mathbf{x}) [\widetilde{K}(\mathbf{x},\mathbf{x})+ \sigma^2_y \mathbf{I}]^{-1} \mathbf{y}\\
    \sigma^* & =  K(x^*_i,x^*_i) - \widetilde{K^*}(x^*_i,\mathbf{x}) [\widetilde{K}(\mathbf{x},\mathbf{x})+ \sigma^2_y \mathbf{I}]^{-1} \widetilde{K^*}(\mathbf{x},x^*_i))
\end{aligned}
\end{equation}
For a given generalized linear function $\Psi(f(x_i^a ))$, we have $\Psi(\mathbf{f}(\mathbf{x}^a )) = \mathbf{W}_{\Psi}^a \mathbf{f}(\mathbf{x}^a) $, where $\mathbf{W}_{\Psi}^a$ is a $N_a \times N_a$ weighting matrix. When $ \mathbf{f}(\mathbf{x}^a) \sim {\cal GP} (0,K(x^a_i, x^a_j)) $, then  $\Psi(\mathbf{f}(\mathbf{x}^a )) = \mathbf{W}_{\Psi}^a \mathbf{f}(\mathbf{x}^a) \sim {\cal GP} (0,\mathbf{W}_{\Psi}^a K(x^a_i, x^a_j)(\mathbf{W}_{\Psi}^a)^T))$ Thus, we have:
\begin{equation}
\begin{aligned}
y^a & = \mathbf{w}^a  [\Psi(\mathbf{f}(\mathbf{x}^a ))]^T\\
& =  \mathbf{w}^a [ \mathbf{W}_{\Psi}^a \mathbf{f}(\mathbf{x}^a) ]^T \sim {\cal GP} ( 0 , \mathbf{w}^a [ \mathbf{W}_{\Psi}^a \mathbf{K}(\mathbf{x}^a,\mathbf{x}^a) (\mathbf{W}_{\Psi}^a)^T] (\mathbf{w}^a)^T)
\end{aligned}
\end{equation}
Following the Eq. \ref{eqn:agg_correlation}, the covariance is given by:
\begin{equation}
    Cov(y^a,y^b) = Cov(\mathbf{w}^a [ \mathbf{W}_{\Psi}^a \mathbf{f}(\mathbf{x}^a) ]^T, \mathbf{w}^b [ \mathbf{W}_{\Psi}^b \mathbf{f}(\mathbf{x}^b) ]^T) \sim {\cal GP} ( 0 , \mathbf{w}^a [ \mathbf{W}_{\Psi}^a \mathbf{K}(\mathbf{x}^a,\mathbf{x}^b) (\mathbf{W}_{\Psi}^b)^T] (\mathbf{w}^b)^T)
\end{equation}
When the function $\Psi(\cdot)$ is a fixed function and $\mathbf{W}_{\Psi}$ are constants we recover the \vbagg model. When the function $\Psi(\cdot)$ is a random function and $\mathbf{W}_{\Psi}$ are Gaussian processes, then $\Psi(\mathbf{f}(\mathbf{x}^a)$ is a GPRN and the aggregation function $y^a$ follows the \cmgpaggr model without composite likelihood corrections.

\end{appendix}

\bibliographystyle{apalike}
\bibliography{main}

\end{document}